\documentclass[lettersize,journal]{IEEEtran}
\usepackage[caption=false,font=normalsize,labelfont=sf,textfont=sf]{subfig}
\usepackage{textcomp}
\usepackage{stfloats}
\usepackage{url}
\usepackage{verbatim}
\usepackage{cite}
\usepackage{times}
\usepackage{epsfig}
\usepackage{graphicx}
\usepackage{amsmath}
\usepackage{amssymb}

\usepackage[pagebackref=true,breaklinks=true,letterpaper=true,colorlinks,bookmarks=false]{hyperref}

\usepackage{booktabs}
\usepackage[export]{adjustbox}
\usepackage[capitalize]{cleveref}
\crefname{section}{Sec.}{Secs.}
\Crefname{section}{Section}{Sections}
\Crefname{table}{Table}{Tables}
\crefname{table}{Tab.}{Tabs.}

\usepackage{multirow}
\usepackage[utf8]{inputenc}
\usepackage{times}
\usepackage[ruled]{algorithm}
\usepackage[noend]{algpseudocode}
\usepackage{enumerate}
\usepackage{todonotes}
\usepackage{mathtools}
\usepackage{tikz-cd}
\usepackage{caption}%,subcaption}

\usepackage{array}
\newcommand{\PreserveBackslash}[1]{\let\temp=\\#1\let\\=\temp}
\newcolumntype{C}[1]{>{\PreserveBackslash\centering}p{#1}}
\newcolumntype{R}[1]{>{\PreserveBackslash\raggedleft}p{#1}}
\newcolumntype{L}[1]{>{\PreserveBackslash\raggedright}p{#1}}

\newcommand{\etal}{\emph{et al}.}

\usepackage[capitalize]{cleveref}
\crefname{section}{Sec.}{Secs.}
\Crefname{section}{Section}{Sections}
\Crefname{table}{Table}{Tables}
\crefname{table}{Tab.}{Tabs.}
\newenvironment{proof}{{\bf Proof:}}{\hfill$\square$}
\newtheorem{remark}{Remark}
\newenvironment{theorem}{{\newline \bf Theorem 1.}}{}

\newcommand{\R}{\mathbb{R}}

\newcommand{\Imm}{\operatorname{Imm}}

\newcommand{\Diff}{\operatorname{Diff}}
\def\argmin{{\operatorname{argmin}}}

\newcommand{\pspace}{\mathcal{T}}
\newcommand{\latentImm}{\mathcal{L}}

\newcommand{\vol}{\operatorname{vol}}
\begin{document}

\title{Basis restricted elastic shape analysis on the space of unregistered surfaces}

%\author{IEEE Publication Technology,~\IEEEmembership{Staff,~IEEE,}
\author{Emmanuel Hartman, Emery Pierson, Martin Bauer, Mohamed Daoudi, Nicolas Charon 
        % <-this % stops a space

\IEEEcompsocitemizethanks{\IEEEcompsocthanksitem
E. Hartman is with Florida State University, Department of Mathematics, E-mail: ehartman@fsu.edu 

E. Pierson  is with the Univ. Lille. E-mail: emery.pierson@courrier.dev 

M. Bauer is with Florida State University, Department of Mathematics, E-mail: bauer@math.fsu.edu

M. Daoudi is with  Univ. Lille, CNRS, Centrale Lille, Institut Mines-Télécom, UMR 9189 CRIStAL, F-59000 Lille, France and IMT Nord Europe, Institut Mines-Télécom, Univ. Lille, Centre for Digital Systems, F-59000 Lille, France, E-mail: mohamed.daoudi@imt-nord-europe.fr

N. Charon  is with University of Houston, Department of Mathematics. E-mail: ncharon@central.uh.edu
}
}

% The paper headers
\markboth{Hartman et. al.: Basis restricted elastic shape analysis on the space of unregistered surfaces}%
{Basis restricted elastic shape analysis on the space of unregistered surfaces}

\maketitle

\begin{abstract}
This paper introduces a new mathematical and numerical framework for surface analysis derived from the general setting of elastic Riemannian metrics on shape spaces. Traditionally, those metrics are defined over the infinite dimensional manifold of immersed surfaces and satisfy specific invariance properties enabling the comparison of surfaces modulo shape preserving transformations such as reparametrizations. The specificity of the approach we develop is to restrict the space of allowable transformations to predefined finite dimensional bases of deformation fields. These are estimated in a data-driven way so as to emulate specific types of surface transformations observed in a training set. The use of such bases allows to simplify the representation of the corresponding shape space to a finite dimensional latent space. However, in sharp contrast with methods involving e.g. mesh autoencoders, the latent space is here equipped with a non-Euclidean Riemannian metric precisely inherited from the family of aforementioned elastic metrics. We demonstrate how this basis restricted model can be then effectively implemented to perform a variety of tasks on surface meshes which, importantly, does not assume these to be pre-registered (i.e. with given point correspondences) or to even have a consistent mesh structure. We specifically validate our approach on human body shape and pose data as well as human face scans, and show how it generally outperforms state-of-the-art methods on problems such as shape registration, interpolation, motion transfer or random pose generation.
\end{abstract}

\section{Introduction}\label{sec:intro}
\noindent {\bf Overview and Main Contributions:}
In this article, we introduce a novel pipeline designed to quantify the geometric difference between the shapes of surfaces. Furthermore, we are not only interested in quantifying shape differences between two individual data points, but we aim to estimate in addition plausible deformation processes between different shapes and allow for statistical shape analysis tasks such as extrapolation of deformations, transposition to new data, and  the generation of random shapes. Finally, the proposed model does not assume a consistent mesh structure across the data, making it applicable to a variety of tasks on surface meshes with real data.

Our approach is grounded in the field of Elastic Shape Analysis (ESA)~\cite{srivastava2016functional} and further leverages the varifold representation of surfaces~\cite{charon2013varifold}, thereby bypassing the common requirement of having consistent mesh structures and available point correspondences across the dataset. 
In contrast to standard ESA, our method relies on enforcing specific structure on the deformation model via the introduction of a data driven basis for the space of admissible shape changes. In the terminology of machine learning, this can be interpreted as a latent space representation but, unlike typical approaches involving autoencoders, in our framework, this latent space is equipped with a non-Euclidean Riemannian metric inherited from the class of second-order invariant Sobolev metrics on the space of surfaces. In comparison to existing geometric deep learning frameworks, our approach's training process is notably straightforward and does not demand a substantial amount of training data. Moreover, as our results suggest, it leads to strong out-of-sample generalization properties when dealing with unseen data. We demonstrate the usability of our framework in a variety of different experiments (registration, interpolation, extrapolation, random shape generation and motion transfer) on two distinct types of data -- human body meshes from the FAUST, DFAUST and SHREC repositories as well as face scans from the COMA dataset.

This work is based on the authors' previous conference publication Hartman~\etal~\cite{Hartman_2023_ICCV}, but introduces several new important additions to that initial approach. This includes in particular:
\begin{itemize}
	\item a more general and in-depth presentation of the mathematical framework of basis restricted elastic shape analysis. Indeed, \cite{Hartman_2023_ICCV} only focused on the setting of human body shapes; here we introduce a general approach applicable to any dataset of surface meshes, see~\cref{sec:model};
    \item a more comprehensive description and justification of the computational model and proposed methodology, including several ablation studies to validate our choice of number of basis vector fields, shape matching functions and Riemannian metric on the latent space, see~\cref{sec:ablation}.
	\item an extended comparison with state-of-the-art latent space methods for body shape analysis (including FARM and 3D-coded), c.f.~\cref{sec:HBResults};
    \item experiments on an extra dataset of human bodies from SHREC, c.f.~\cref{sec:HBResults}.
	\item a new application of the method on different data highlighting the effectiveness of this approach for scans of human faces c.f. ~\cref{ssec:coma}.
 \item an open source coding package for basis restricted ESA with precomputed bases for human body and face analysis, available at \url{https://github.com/emmanuel-hartman/BaRe-ESA}.
\end{itemize}

\subsection{Related Work and Motivation}
Analyzing three-dimensional ($3$D) \emph{surfaces} has become an increasingly important topic, where the need for such algorithms is motivated by the emergence of high-accuracy $3$D scanning devices, that have resulted  in a significant increase in the availability of such data. The resulting application range from human health analysis~\cite{DesrosiersIVC2017}, facial animation ~\cite{QinSigraph2023,OtberdoutCVPR2022}, computer graphics ~\cite{DengCGF2022} or synthetic human data generation ~\cite{ZhangCvpr2020} to computational anatomy \cite{grenander1998computational}.

Although the framework developed in this article is fairly general and, we believe, could be relevant for a variety of real data applications, our simulations will primarily focus on datasets of 3D human bodies and faces. These involve particularly challenging problems given the high degree of variation in shape and pose, and the lack of point correspondences and consistent mesh structure across such datasets. 

\noindent {\bf Geometric shape analysis:} The general field of Riemannian shape analysis has produced several mathematical frameworks and numerical pipelines to tackle some of the key problems in the comparison and statistical analysis of $3$D surfaces. These models are built from a Riemannian metric on a "shape space", in which the "shape" of a surface is usually regarded as what information remains after factoring out shape-preserving transformation groups such as reparametrizations or rigid motions. Two main frameworks have in particular stood out in constructing Riemannian metrics on such shape spaces: on one hand, the diffeomorphic approach of \cite{beg2005computing,younes2019shapes} and, on the other, the elastic metric setting introduced in \cite{younes1998computable,srivastava2016functional}. An important aspect in both models is that the formulation of basic shape analysis tasks such as the estimation of the geodesic distance between two given surfaces for instance, is typically framed as the minimization of a \textit{reparametrization invariant matching energy} in which computation of the distance and of the optimal registration (i.e. of the unknown point correspondences) must be tackled \textit{jointly}. This should be viewed in sharp contrast with the majority of traditional approaches in shape analysis~\cite{audette2000algorithmic} in which registration is performed as a pre-processing step using methods such as functional maps~\cite{ovsjanikov2012functional} and where the subsequent analysis is then done independently of this registration. This practice has been, however, increasingly questioned as it can, in some cases, lead to a severe loss of data structure/information or generate bias in the analysis, see e.g.~\cite{srivastava2016functional} and the references therein. On the other hand, the joint estimation of distance and registration often induces several practical challenges in particular when working with simplicial meshes such as triangulated surfaces. Some approaches~\cite{Kurtek_2012_PAMI,jermyn2017elastic, Su_2020_CVPRW, tumpach_gauge_2016,laga20224d} rely on analytical representations or approximations for the surfaces and the reparametrization group (using e.g. spherical harmonics) but are therefore often limited to a predefined topology. As an alternative, it was proposed, first for the diffeomorphic model in \cite{vaillant2005surface,charon2013varifold}, and later adapted to the ESA framework \cite{Bauer_2021_IJCV, hartman2023elastic}, to instead introduce discrepancy loss functions built from measure representations of surfaces. Those discrepancy functions, in particular the metrics derived from the framework of \textit{varifolds}, have been shown to provide robustness to scan inconsistencies, such as varying mesh samplings and topological noise.

Despite those successes, one of the key remaining limitation of Riemannian shape analysis is the fact that pure geodesic trajectories are often not inherently representative of realistic longitudinal changes in real data. For instance, in one of the data application of this paper, it has been observed that simple geodesic interpolation between two human body poses does not generally reproduce the natural body motion that would be expected, c.f.~Section~VI of the supplementary material. An important current research challenge is thus to develop ways to enforce various types of physical, biological or data-specific constraints within Riemannian shape frameworks. In the diffeomorphic setting, some progress has been made towards that goal either through the introduction of sub-Riemannian \cite{arguillere14:_shape,gris2018sub} or other types of constrained models \cite{charlier2018distortion,hsieh2022diffeomorphic,charon2023shape}. Yet these approaches are typically built around user specified constraints or principles rather than being entirely data-driven and are also known to be numerically costly when working with high resolution data. The basis restricted approach of the present work in part overcomes those difficulties by leveraging, on the one hand, the advantages of the elastic metric framework when it comes to numerical complexity and, on the other hand, by extracting from the dataset itself the adequate constrained subspace of deformations. In the registered setting a similar approach has been used in the conference paper~\cite{Pierson_2022_WACV} by some of the authors and Tumpach. Such basis models are highly related to latent space models, popular in geometric deep learning~\cite{bronstein2017geometric,bronstein2021geometric}, which we will describe next.

\noindent{\bf Low dimensional deep deformation models:}
Recently, deep deformation models have become increasingly popular for shape representation and deformation. By reproducing the results of Generative Adversarial Networks~\cite{NIPS2014_5ca3e9b1} or Variational Auto Encoders~\cite{KingmaW13} on deformable shapes, geometric deep learning~\cite{bronstein2017geometric} methods have shown that deforming shapes in their latent representation can be efficient. These approaches propose to build a deformation model for different types of deformable shapes, such as the human body~\cite{bouritsas2019neural, lemeunier2022representation, huang2021arapreg, groueix2018b}, the human face~\cite{bouritsas2019neural, otberdout2022sparse}, or animals~\cite{huang2021arapreg} based on a limited training set of parameterized shapes. 

However, those methods need to deal with parameterization invariance at inference. This is often done using a PointNet encoder~\cite{Qi_17_CVPR}, which maps a shape to its latent vector independently of its discretization. Other approaches have been proposed, but they come with an high training cost~\cite{TrappoliniNips2021} or use intrinsic quantities such as the Laplacian~\cite{sharp2022diffusionnet, wiersma2022deltaconv}, that can be sensitive to topological changes. We note, however, that in practice, the invariance of those methods remains limited, because of  their reliance on large datasets of parameterized surfaces for training purposes. They often need additional post-processing registration steps in inference to reproduce plausible geometric reconstruction of shapes~\cite{huang2021arapreg, groueix20183d, TrappoliniNips2021}.

Moreover, the performance of these methods is often limited in the context of large deformations: they regularly fail to sufficiently learn the non-linear map from the flat latent space to the shape space. Consequently, they are lacking generalizability when confronted with unseen data. To address these issues multiple deformation energy losses have been introduced in the training phase, such as geodesic distances~\cite{cosmo2020limp}, ARAP~\cite{huang2021arapreg, GLASSCvpr2022}, or volumetric constraints~\cite{atzmon2021augmenting}. Manifold regularization of learned pose spaces~\cite{10.1007/978-3-031-20065-6_33, freifeld2012lie} has also been proposed. Those geometric quantities however increase the total training costs of those approaches. 

In contrast, our approach does not rely on a non-linear map but imposes an affine map, called the affine decoder, from a given low dimensional latent space to a corresponding space of shapes. This space is defined using pre-estimated basis. We impose non-linearity on the deformation space via the pullback of a second-order, parametrization-invariant, Sobolev (Riemannian) metric. The registration of a scan becomes an interpolation problem between the template and the scan representation in the low dimensional space~\ref{eq:matching}, proposing plausible registrations of the shape. Moreover, interpolation and extrapolation problems are formulated as geodesic boundary and initial value problems and are easily implemented using modern scientific computation libraries.

\section{Mathematical background}\label{sec:model}
\subsection{The Riemannian shape space of immersed surfaces}
In this article, the "shapes" of interest are surfaces immersed in the Euclidean space $\mathbb{R}^3$. Mathematically, and from the continuous viewpoint, we define a \textit{parametrized shape} as an immersion from a generic 2D parameter domain $\pspace$ (a compact 2-dimensional manifold) into $\mathbb{R}^3$, i.e. a smooth mapping $q:\pspace \rightarrow \mathbb{R}^3$ such that the Jacobian map $dq(u)$ is injective for all $u \in \pspace$. For instance, $\pspace$ can be taken as a compact domain of $\mathbb{R}^2$ if one considers open surfaces (such as human faces) or the sphere $S^2$ in the case of closed surfaces (such as whole human body surfaces). We shall denote by $\operatorname{Imm}$ the space of all immersions from $\pspace$ to $\mathbb{R}^3$.        

\begin{figure}
    \includegraphics[width=0.45\textwidth]{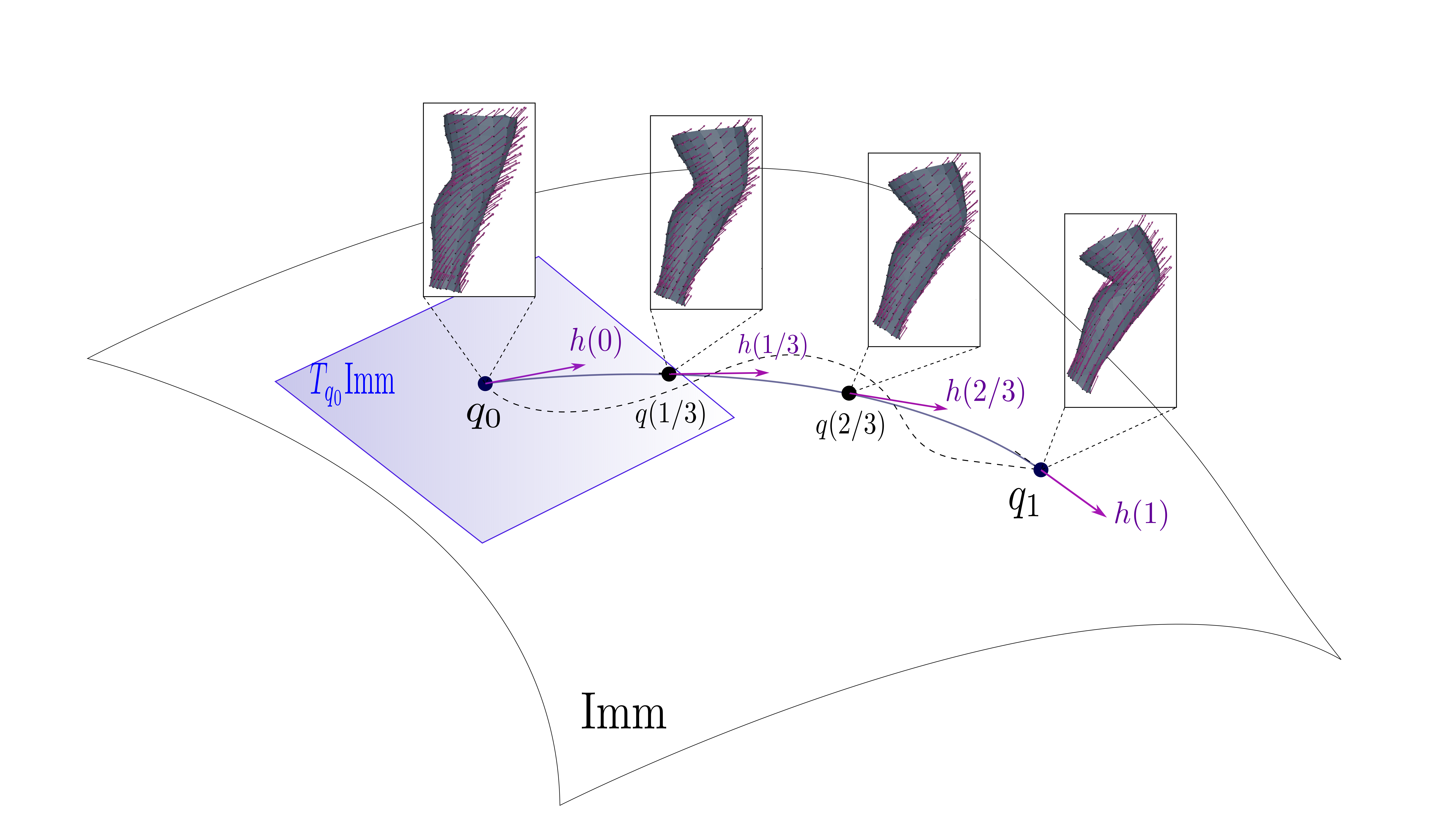}
    \caption{Illustration of the Riemannian shape space of parametrized immersed surfaces.} \label{fig:imm_space}
\end{figure}

In order to provide a quantitative way to compare such shapes, one needs to introduce a similarity measure on $\operatorname{Imm}$. As pointed out in the introduction, we are here interested in similarity measures that originate from a Riemannian setting, in other words we wish to view $\operatorname{Imm}$ as an infinite dimensional manifold and equip it with a Riemannian metric. In this setup the corresponding geodesic distance function provides the similarity measure for shape comparison. In addition this will allow us to reduce tasks such as shape interpolation and extrapolation to geometric operations -- the geodesic initial and boundary value problem. First, note that, as an open subset of $C^{\infty}(\pspace,\mathbb R^3)$, $\operatorname{Imm}$ can be directly viewed as a Fr\'echet manifold for which the tangent space $T_q \operatorname{Imm}$ at each immersion $q$ can be naturally identified with $C^{\infty}(\pspace,\mathbb R^3)$ or equivalently as the space of smooth deformation fields along the parametrized surface $q$, cf. Figure~\ref{fig:imm_space} for an explanatory illustration of the shape space of parametrized immersed surfaces. 

In this setting, a Riemannian metric $G$ is a family of inner products 
$G_q:C^{\infty}(\pspace,\mathbb R^3)\times C^{\infty}(\pspace,\mathbb R^3)\to \mathbb R_+ $
that depends smoothly on the foot point $q\in \Imm$. We further recall that, from $G$, one obtains a "distance" on $\Imm$ which, for any $q_0,q_1 \in \Imm$, is obtained as:
\begin{equation}
\label{eq:param_distance}
    d_G(q_0,q_1)^2 = \inf \left\{\int_0^1 G_{q(t)}(h(t)), h(t))  \right\} 
\end{equation}
where $h(t) = \partial_t q(t) \in T_{q(t)} \Imm$ and where the infimum is taken over all paths $q:[0,1] \rightarrow \Imm$ with $q(0) = q_0$ and $q(1)=q_1$. We call \eqref{eq:param_distance} the \textbf{parametrized matching problem}. Any minimizing path, if it exists, is then a \textit{geodesic} between $q_0$ and $q_1$. 

To define the Riemannian metric $G$ we will rely on the setting of elastic shape analysis (ESA) which has derived various families of metrics that further satisfy the key property of reparametrization-invariance. To explain this property we introduce the notion of a \textit{reparametrization} $\phi$ as an element of the diffeomorphism group $\Diff(\pspace)$, i.e., the set of smooth and bijective maps on the parameter domain $\space$. This group  acts on any given immersion $q$ by right composition, i.e.,  $(q,\phi)\mapsto q \circ \phi$. The metric $G$ is called reparametrization invariant if for any $\phi \in \Diff(\pspace)$ we have
\begin{equation}
   G_{q \circ \phi}(h\circ \phi,k\circ \phi) = G_q(h,k)
\end{equation}
and the importance of this property will become clear below in Section~\ref{ssec:param_vs_unparam}, where we will quotient out the action of this group. 

Perhaps the simplest of those metrics is the so called invariant $L^2$ metric defined for all $q \in \Imm$ and $h,k \in C^{\infty}(\pspace,\mathbb R^3)$ via:
\begin{equation*}
  G_q(h,k) = \int_{\pspace} \langle h , k \rangle \vol_q  
\end{equation*}
where $\vol_q$ is the volume measure on $\pspace$ induced by $q$, that is, denoting $(u_1,u_2)$ some coordinates on $\mathcal{T}$, $\vol_q = |\partial_{u_1} q \wedge \partial_{u_2} q|$. The integration with respect to this induced volume measure is precisely what leads to the invariance of the metric (and by extension of the geodesic distance).  One crucial shortcoming of the above metric however, which was first shown by Michor and Mumford in \cite{michor2005vanishing,bauer2012vanishing}, is that the associated $d_G$ turns out to be fully degenerate and a fortiori not a true distance, i.e., with respect to this metric all shapes are considered to be equal. 

One way to address this issue is by introducing higher-order metrics on $\Imm$. In this article, we shall  focus on the class of \textit{second order invariant Sobolev metrics} which has shown several desirable properties in past works \cite{bauer2011sobolev,hartman2023elastic}. More specifically we will consider the 6-parameters family of metrics obtained by the following combination of 0-th, 1-st and 2-nd order terms weighted by the nonnegative constants $a_0,a_1,b_1,c_1, d_1$ and $a_2$: 
\begin{equation}
\begin{aligned}
&G_q(h,k)=\int_{\pspace}\bigg( a_0 \langle h,k \rangle +
a_1 g_q^{-1}(dh_m,dk_m)\\&\qquad\qquad +b_1g_q^{-1}(dh_+,dk_+)+ c_1g_q^{-1}(dh_\bot,dk_\bot)\\&\qquad\qquad+ d_1 g_q^{-1}(dh_0,dk_0)
+a_2 \langle\Delta_q h,\Delta_q k\rangle\bigg)\vol_q.
\end{aligned}
\label{eq:Sobolev_metric}
\end{equation}
In the above, $dh$ denotes the vector-valued 1-form on $\pspace$ given by the differential of $h$ which we can alternatively view, in a given coordinate system, as a $3 \times 2$ matrix field on $\pspace$, $g_q$ is the pullback of the Euclidean metric on $\R^3$ which we may view as a $2 \times 2$ symmetric positive definite matrix field on $\pspace$, in which case $g_q^{-1}(dh,dh) = \operatorname{tr}(dh g_q^{-1} dh^T)$. The second-order term involves the vector Laplacian $\Delta_q$ induced by the parametrization which in coordinates can be written as $\Delta_q h = \frac{1}{\sqrt{\det(g_q)}}\partial_{u_1}\left(\sqrt{\det(g_q)}g_q^{u_1 u_2}\partial_{u_2} h \right)$. Lastly, let us briefly comment on the particular splitting of the first-order part of the metric in the four different terms appearing in \eqref{eq:Sobolev_metric}. For the sake concision, we shall refer the interested reader to the appendix (or to \cite{su2020shape,charon2023shape}) for the technical definition of the orthogonal decomposition of $dh$ into the sum of the tensors $dh_m, dh_+, dh_\bot, dh_0$. We will only say at this point that such a splitting is motivated by its interpretation from linear elasticity theory, with the terms weighted by $a_1,b_1,c_1$ in \eqref{eq:Sobolev_metric} corresponding to thin shell shearing, stretching and bending energies induced by the deformation field $h$ respectively.

Consequently, the class of invariant $H^2$ metrics \eqref{eq:Sobolev_metric} provides a flexible family allowing, through the selection of the weighting coefficients, to emphasize or penalize different types of deformations. 
Each of those metrics is reparametrization-invariant and unlike the $L^2$ case
it induces a true distance on $\Imm$:
\begin{theorem}
\emph{Let $a_0>0$ and let either $a_1,b_1,c_1,d_1>0$ or $a_2>0$ then the induced geodesic distance of the metric $G$ on the space $\Imm$ is non-degenerate, i.e., for any two surfaces $q_0,q_1\in \Imm$ with $q_0\neq q_1$ we have}
$
d_G(q_0,q_1)>0
$.\end{theorem}

For a proof of this result we refer to the supplementary material. Furthermore, as we shall explain later, there are natural discretization schemes to compute such second-order metrics on e.g. triangulated meshes.

\subsection{Quotienting out reparametrizations}
\label{ssec:param_vs_unparam}
Note that the model described so far leads to distances and geodesics between parametrized shapes. From a practical standpoint, this intrinsically assumes known point to point correspondences, namely point $q_0(u)$ on the source surface is matched to $q_1(u)$ on the target. Apart from pre-registered datasets (such as the DFAUST one described below), it is common in most applications that raw or segmented surface meshes do not come with such given correspondences, and even display inconsistent number of vertices and/or mesh structures. Thus one is typically interested in comparing surfaces \textbf{independently} of how they are parametrized/sampled.

Mathematically, this can be done by looking at the quotient shape space $\mathcal{S}=\Imm / \Diff(\pspace)$ of the equivalence classes $[q] = \{q \circ \phi \ : \ \phi \in \Diff(\pspace)\}$ of all possible reparametrizations of $q$. A key advantage of the invariant metric framework introduced in the previous section, and in particular of the invariant Sobolev metrics \eqref{eq:Sobolev_metric}, is that one can recover a Riemannian distance on $\mathcal{S}$ as follows. Given unparametrized surfaces $[q_0]$ and $[q_1]$, the quotient distance is obtained by fixing a parametrization $q_0$ and solving the following \textbf{unparametrized matching problem}:
\begin{equation}
\label{eq:unparam_distance}
    \overline{d}_G([q_0],[q_1])^2 = \inf_{(q(\cdot),\phi)} \left\{\int_0^1 G_{q(t)}(\partial_t q(t), \partial_t q(t)) \right\} 
\end{equation}
where the minimization is now over paths $q:[0,1] \rightarrow \Imm$ \textbf{and} reparametrization $\phi \in \Diff(\pspace)$, with the constraint that $q(0)=q_0$ and $q(1) = q_1 \circ \phi$ i.e. $[q(1)]=[q_1]$. In other words, the quotient distance is obtained by jointly finding an optimal path from $q_0$ to an optimal reparametrization of the target. 

However, the variational problem \eqref{eq:unparam_distance} is generally challenging to tackle and to implement on discrete surface meshes. It involves estimating parametrizations of the two surfaces over a predefined domain (such as the sphere) and then requires discretizing and optimizing over the group $\Diff(\pspace)$ \cite{jermyn2017elastic,su2020shape}. An alternative approach in registration problems is rather to enforce the matching constraint $[q(1)]=[q_1]$ indirectly via a discrepancy function $\Gamma([q(1)],[q_1])$ that only depends on the unparametrized shapes and therefore consider the \textbf{relaxed matching problem}: 
\begin{equation}
\label{eq:matching}
   \inf_{q(\cdot)} \left\{\int_0^1 G_{q(t)}(\partial_t q(t), \partial_t q(t)) + \lambda \Gamma([q(1)],[q_1]) \right\}  
\end{equation}
in which $\lambda >0$ acts as a Lagrange multiplier for the terminal constraint, and the minimization is now only over parametrized surface paths $t \mapsto q(t)$; in other words, we bypass the need for directly optimizing reparametrizations. 

To define $\Gamma$, one typically introduces a measure of similarity between the geometric point sets $q_1(\pspace)$ and $q(1)(\pspace)$; we discuss a few possible options in the Supplementary Material, including the Hausdorff and Chamfer distances often used for that purpose in computer vision. In this paper, following many other works and our previous publication \cite{hartman2023elastic}, we instead rely on similarity terms derived from geometric measure theory, specifically the family of kernel metrics on the space of \textit{varifolds} \cite{kaltenmark2017general}. A notable advantage of this framework is that it leads to actual distances that can be differentiated with respect to the point positions of either shape. Although we will abstain from presenting this construction in the main text for concision, we refer the reader to the Appendix for details and qualitative comparison of varifold metrics with other classical point set discrepancies.

\section{Restricted latent space model}
\label{sec:latent_space}
As highlighted in the introduction, one limitation of the general $H^2$ metric framework is that it does not impose any restriction on deformation fields beyond the energy penalties in the metric \eqref{eq:Sobolev_metric}. When it comes to modelling human body motion for example, it has been observed that geodesics between two poses most often do not emulate a "natural" interpolation of the pose, despite the flexibility in the choice of metric coefficients. A second practical downside is the numerical complexity of having to solve a very high dimensional optimization problem over paths of surfaces for any estimation of a distance and geodesic, which can become quite significant when generalizing that approach for more complex statistical tasks such as Fr\'{e}chet mean estimation or parallel transport.  

\subsection{Latent space representation} 
As a way to address the above challenges, we propose a simplified and \textbf{linearized finite dimensional} shape space model by restricting ourselves to parametrized surfaces $q$ that result from a fixed template surface $\bar{q} \in \Imm$ and a predefined admissible set of $P$ linearly independent deformation fields $\{h_i\}_{i=1}^{P}$ of the template. More precisely, with the affine mapping $F:\R^P \rightarrow C^\infty(\pspace,\R^3)$ defined by:
\begin{equation}
 F: (\alpha_i)_{i=1,\ldots,P} \mapsto \bar q + \sum_{i=1}^{P} \alpha_i h_{i},
\end{equation}
we introduce the space $\latentImm_{\bar{q}} = F(\R^P) \bigcap \Imm$.

Any surface $q \in \latentImm_{\bar{q}}$ can be then represented uniquely by a finite-dimensional vector $\alpha=(\alpha_i) \in \R^P$ we will call the \textit{latent code} of $q$, thus allowing us to work on a potentially much lower-dimensional space.  Yet $\latentImm_{\bar{q}}$ should still remain rich enough so as to express the predominant geometric variations in the dataset of interest. As we shall address in~\Cref{sec:basis}, this suggests using a basis $\{h_i\}$ that is built in a \textit{data-driven way}. Furthermore, this latent space model allows the use of composite bases, where different subsets of vector fields are associated to distinct types of morphological variations. This will prove particularly relevant to the applications of this paper when we are interested in e.g. disentangling body pose from body type changes or facial expression from facial morphology changes.
\begin{remark}
    Note that, in general, $\latentImm_{\bar{q}}$ is an open subset of the affine space $F(\R^P)$ and contains $\bar{q}$. However, not all elements of $F(\R^P)$ are immersions unless certain specific conditions on the vector fields are satisfied. This holds in particular if for all $i=1,\ldots,N$ and $u\in \pspace$, $dh_i(u)^T d\bar{q}(u) =0$. However, we will not assume this condition in the rest of the paper.
\end{remark}

\subsection{Induced Riemannian metric}
The next logical question to address is which metric to take on the above latent space. In sharp contrast to most encoder models in geometric deep learning which often implicitly consider the standard Euclidean structure, our approach is rather to take advantage of the properties of invariant metrics on shape spaces and pull the metric back to the latent space. Namely, for any $\alpha \in F^{-1}(\latentImm_{\bar{q}})$ and $\beta,\eta \in \R^P$, let us define:
%\begin{equation*}
    $\overline{G}_{\alpha}(\beta,\eta):= G_{F(\alpha)}(d_\alpha F(\beta ),d_\alpha F(\eta ))$
%\end{equation*}
in which $G$ is a Riemannian metric on $\Imm$ which we shall take from the invariant family \eqref{eq:Sobolev_metric}. As the mapping $F$ is affine, this pull back metric on $\R^P$ can be expressed more explicitly as:
\begin{equation*}
    \overline{G}_{\alpha}(\beta,\eta)= G_{F(\alpha)}\left(\sum_{i=1}^P \beta_i h_i , \sum_{j=1}^P \eta_j h_j \right) = \beta^T \overline{G}_{\alpha} \eta
\end{equation*}
where, in the last equation, $\overline{G}_\alpha = [G_{F(\alpha)}(h_i,h_j)]$ is the symmetric positive definite $P \times P$ matrix giving the metric at the latent code $\alpha$. Estimation of the distance between any two surfaces $q_0=F(\alpha_0)$ and $q_1=F(\alpha_1)$ then reduces to standard finite-dimensional Riemannian geometry and is obtained by finding a path of coefficients $t \mapsto \alpha(t) \in \R^P$ minimizing $E(\alpha) = \int_0^1 (\partial_t \alpha)^T \overline{G}_\alpha (\partial_t \alpha) dt$ with $\alpha(0) = \alpha_0$ and $\alpha(1) = \alpha_1$.

\section{Shape analysis in latent space}
Relying on the latent space representation and its Riemmanian metric introduced in the previous section, one can perform efficiently a variety of shape analysis related tasks, which we describe in the following paragraphs.  

\subsection{Calculating latent space representations}\label{sec:latentcodes}
We start by describing how we can calculate a latent space representation that is (up to numerical accuracy) independent of the parametrization of the surface, i.e.,  given a surface $q\in \Imm$  we aim to find a latent code representation $\alpha$ such that $F(\alpha)= q\circ\phi$ for some (unknown) reparametrization function $\phi\in \Diff(\pspace)$. To tackle this problem  we rely again on the varifold similarity term, i.e., we reformulate the latent representation problem as the task of finding a latent code representation $\alpha$ such that 
 \begin{equation}\label{eq:latent_code1}
 \Gamma(F(\alpha),q)=0. 
 \end{equation}
 One remaining difficulty is that, for most datasets such as those of Section \ref{sec:datasets}, raw surface scans are not given with consistent mesh structures and a fortiori cannot be assumed to all belong to $\latentImm_{\bar{q}}$ for a given fixed template $\bar{q}$. To circumvent this difficulty we consider  a relaxed formulation of the latent code representation problem; instead of searching for a latent code $\alpha$ satisfying equation~\eqref{eq:latent_code1} we simply aim to minimize the varifold distance $\Gamma(F(\alpha),q)$ over all latent codes $\alpha\in \mathbb R^P$. 
In our experiments it turned out to be beneficial to add an extra regularizing term to this minimization problem, which we choose to be the geodesic distance of $F(\alpha)$ to the template $\bar q$, i.e., we minimize the energy
\begin{equation}
\left(\Gamma(F(\alpha),q)+ \frac1\lambda d^{\latentImm_{\bar{q}}}_G(\bar{q},F(\alpha))^2\right)\label{eq:latent_code3}
\end{equation} 
over all $\alpha\in \R^P$, where $\lambda>0$ is a weight parameter. 
Using the definition of the geodesic distance $d^{\latentImm_{\bar{q}}}_G$ on the latent space $\latentImm_{\bar{q}}$ this requires us to minimize the path energy 
\begin{equation}
  \Gamma(F(\alpha(1)),q)+\frac1\lambda \int_0^1 \overline{G}_{\alpha}(\partial_t\alpha(t),\partial_t\alpha(t)) dt  \label{eq:latent_code3}
\end{equation} 
over all paths $\alpha: [0,1]\to \R^P$. Numerically, we consider time-discrete paths of coefficients $\alpha=(\alpha(0), \alpha(1/T),\alpha(2/T),\ldots,\alpha(1))$ for a selected number of time steps $T$, with $\partial_t \alpha$ being approximated by forward finite difference. Furthermore, $q$ and $\bar{q}$ are in practice given as sets of vertices and triangular meshes while each $h_i$ is of a collection of vectors sampled on the vertices of $\bar{q}$. This turns the problem into an unconstrained minimization over $\R^{P(T-1)}$ for which we use the L-BFGS algorithm of the \textit{scipy} library; here the free variables are only  in $\R^{P(T-1)}$ as the path starts at $\bar q$ and thus $\alpha(0)=0$. The precise discretization of the different terms in \eqref{eq:sym_match_energy}, based on the principles of discrete differential geometry, is detailed in the Supplementary Material. Our implementation, that builds on some of the authors' previous package for surface matching\footnote{\url{https://github.com/emmanuel-hartman/H2_SurfaceMatch}}, is done in Python and relies on libraries such as \textit{PyTorch} and \textit{PyKeops} which allow to automatically differentiate those terms on the GPU.  Our implementation is also publicly available on Github\footnote{\url{https://github.com/emmanuel-hartman/BaRe-ESA}} and relies on the same libraries.

\subsection{Shape comparison and interpolation}
Quantifying the global difference between surfaces is generally essential when attempting for example to cluster data in a population. The Riemannian metric setting gives a direct way to measure such differences via the distance itself and, what is more, lead to geodesic paths that interpolate between the objects. The availability of such geodesic paths has the double advantage of allowing to interpret the properties and behaviour of the distance while also providing a way to reconstruct a dynamical evolution from one data point to another. 

Within the framework of Section \ref{sec:latent_space}, we have seen that the estimation of distance and geodesics between two surfaces $q_0=F(\alpha_0)$ and $q_1=F(\alpha_1)$ in $\mathcal{L}$ can be done by finding a path of least Riemannian energy in the latent space, i.e., by minimizing the path energy
\begin{equation}
 \int_0^1 \overline{G}_{\alpha}(\partial_t\alpha(t),\partial_t\alpha(t)) dt  \label{eq:geodesic_latent_space}
\end{equation} 
over all paths $\alpha: [0,1]\to \R^P$ such that $\alpha(0)=\alpha_0$ and $\alpha(1)=\alpha_1$. Discretizing the path in time $t$ this leads to an unconstrained minimization problem over $\R^{P(T-2)}$ with the free variables being $\alpha(1/T),\alpha(2/T),\ldots,\alpha((T-1)/T))$ as $\alpha(0)=\alpha_0$ and $\alpha(1)=\alpha_1$ are fixed.

Given new data, for which we have not yet calculated a latent space representation, we could proceed as follows: calculate first a latent space representation using the method of the previous section and then solve the geodesic 
problem using the above algorithm. In practice it is, however, more effective to solve both of these tasks in one step. This can be done using again the varifold distance and by considering the path minimization problem

\begin{multline}
 \int_0^1 \overline{G}_{\alpha}(\partial_t\alpha,\partial_t\alpha) dt\\+\lambda \Gamma(F(\alpha(0)),q_0)+\lambda \Gamma(F(\alpha(1)),q_1). \label{eq:sym_match_energy}
\end{multline}    
where $\alpha: [0,1]\to \R^P$ is again a path in the latent space $\latentImm_{\bar{q}}$. The presence of the two discrepancy terms in \eqref{eq:sym_match_energy} is necessary to make the above problem well-defined for \textbf{any} $q_0$ and $q_1$ in $\Imm$ and not just in $\latentImm_{\bar{q}}$. The solution \eqref{eq:sym_match_energy} can be thus interpreted as the distance and geodesic between the closest approximations of $q_0$ and $q_1$ by elements of the latent space.

\subsection{Shape extrapolation}
The shape extrapolation problem consists in predicting the future evolution of a surface given an initial deformation direction. In our Riemannian framework this reduces to solving the geodesic equation  with given initial condition $q(0)=q_0$ (the initial pose) and $\partial_t q(0)=h$ (the deformation direction), cf. Figure~\ref{fig:imm_space}. The geodesic equation is the first order optimality condition of the energy functional; it is a non-linear PDE, that is second order in time $t$ and forth order in space (twice the order of the metric). For the exact formula of this equation, which is rather lengthy and not particularly insightful, we refer the interested reader to the literature, see eg.~\cite{bauer2011sobolev}. To solve such initial value problems in our latent space, we modify methods of discrete geodesic calculus~\cite{rumpf2013discrete} to our setting. We approximate the geodesic starting at $\alpha^0$ in the direction of $\beta$ with a PL path with $N+1$ evenly spaced breakpoints. At the first step, we set $\alpha^1=\alpha^0+\frac{1}{N}\beta$ and find $\alpha^2$ such that $F(\alpha^1)$ is the geodesic midpoint of  $F(\alpha^0)$ and  $F(\alpha^2)$, i.e., we solve for $\alpha^2$ such that 
\begin{equation*}
\alpha^1 = \underset{\Tilde{\alpha}}{\argmin} [\overline{G}_{\alpha^0}(\beta_0,\beta_0)+\overline{G}_{\Tilde{\alpha}}(\Tilde{\beta},\Tilde{\beta})]
\end{equation*} where $\beta_0=\Tilde{\alpha}-\alpha^0$ and $\Tilde{\beta}=\alpha^2-\Tilde{\alpha}$.
Differentiating with respect to $\Tilde{\alpha}$ and evaluating the resulting expression at $\alpha^1$, we obtain the system of equations 
\begin{multline}
    2\overline{G}_{\alpha^0}(\beta_0,h_i)-2\overline{G}_{\alpha^1}(\Tilde{\beta},h_i)+ D_{\alpha^1}\overline{G}_{\cdot}(\Tilde{\beta},\Tilde{\beta})_i=0, \\
    2\overline{G}_{\alpha^0}(\beta_0,k_i)-2\overline{G}_{\alpha^1}(\Tilde{\beta},k_i)+ D_{\alpha^1}\overline{G}_{\cdot}(\Tilde{\beta},\Tilde{\beta})_{i+m}=0  \label{eq:first_order_condition_ivp}
\end{multline}
where $\{h_i,k_i\}$ is our basis of deformations as introduced above. We denote the system of equations in~\eqref{eq:first_order_condition_ivp} by $\Phi(\alpha^2; \alpha^1, \alpha^0) = 0$, where we stress again that $\alpha^0$ and $\alpha^1$ are here fixed and known. We solve this system of equations for $\alpha^2$ using a nonlinear least squares approach, i.e., by computing
\begin{equation*}
    \alpha^2 = \underset{\Tilde{\alpha}}{\argmin} \| \Phi(\Tilde{\alpha}; \alpha^1, \alpha^0) \|_2^2.
\end{equation*}
We repeat this process $N-1$ times, thereby constructing the discrete solution up to time $t=1$.

\subsection{Motion transfer in latent space}
As previously discussed, composite bases offer a means to independently depict various modes of shape deformation. Specifically, when applied to human body and facial morphology, these bases allow us to separate identity and pose alterations, enabling motion transfer. In practical terms, when presented with a series of unregistered scans depicting a single identity engaged in an action, we can obtain latent code representations for each frame of the action. We then substitute the coefficients of the shape basis with the shape coefficients of the desired identity. This process yields a sequence of shapes that faithfully embodies the desired motion transferred onto the desired identity. Note that this is significantly simpler (albeit different) than performing parallel transport in the Riemannian manifold of surfaces as done in e.g. \cite{hartman2023elastic}.

\subsection{Random shape generation}
Additionally, we can utilize the Riemannian structure on our latent space representation to offer a data-driven method for generating random shapes from unregistered data. We may do this by learning an empirical distribution on the tangent space of the template shape. Given a data set of unregistered shapes, we solve the latent code retrieval problem and compute the initial vectors of the resulting geodesics in the latent space. We can then fit Gaussian mixture model on the resulting collection of tangent vectors and solve the initial value problem from the template in the direction of the vector generated from this model. In the case where we compute multiple bases to describe different modalities of shape change, the model may be fit to independently generate different types of shape changes.

\vskip2ex

\section{Experimental Methodology}
In this section we will describe the different datasets, which we will use in the experimental section, the corresponding basis construction and the choice of parameters. In addition we will present different ablation studies, that further motivate the chosen energy functional.
\subsection{Used Datasets}\label{sec:datasets}
\noindent{\bf Human Body datasets:}
The main type of data considered in this article consists of human body scans. To construct our basis we will make use of the publicly available 
Dynamic FAUST (DFAUST)~\cite{dfaust:CVPR:2017} dataset. 
This dataset contains high quality scans, along with corresponding registered meshes that will be used as training data. More specifically
DFAUST~\cite{dfaust:CVPR:2017} is comprised of 4D scans captured at 60 Hz of 10 individuals performing 14 in-place motions. Due to the high speed of the recording, DFAUST scans contains several singularities in the surface, such as holes or even artificial objects (eg. parts of walls). The corresponding registered surfaces to each scan are created using image texture information and a novel body motion model. A set of 7 long range sequence are left for testing. The remaining 133 sequences, which we denote DFaustT, make up the training set from which we compute the deformation and motion basis.

For the quantitative experiments, we consider three testing datasets on which we validate our model trained on DFaustT: First, we consider a subset of the static FAUST dataset~\cite{Bogo:CVPR:2014} for testing our models performance for registration and point correspondences. The static FAUST dataset is a 3D static scan data set designed for human mesh registration tasks, that contains scans of minimally clothed humans and corresponding registered meshes. We selected scans of 10 individuals in 9 different poses from the training set that show no rotations along with the corresponding ground truth registrations and use them as our first testing set, denoted FaustE. Secondly, we consider a subset of the SHREC dataset~\cite{shrec19} to demonstrate the generalizability of our model in shape reconstruction tasks. This dataset, denoted SHREC, contains human shapes from significantly different modalities than that of our training set including scans of clothed humans and synthetic shapes of human bodies. For our third and final testing set, we divide the 7 sequences from DFAUST set aside for testing into 10 representative mini-sequences which we use to evaluate our framework's ability to reconstruct human motions. We denote this DFaustE.

\noindent{\bf Face scan datasets:}
As a secondary type of data, we consider human face scans from the COMA~\cite{ranjan2018generating} dataset. 
This dataset contains high-quality scans of human faces, along with corresponding registered meshes in the FLAME topology~\cite{li2017learning} that will be used as training data. More specifically COMA is comprised of 4D scans of human faces captured at 60 Hz of 12 individuals performing 12 extreme facial expressions. The scans are available as raw scans of the whole face and often contain significant parts of the chest that are not present in the final registrations. Moreover, some detailed parts can be cropped or disappear in the scans, e.g. ears of the individual. The corresponding registered surfaces to each scan are created using image texture information, face landmarks and the FLAME model. 
We provide qualitative results, on this dataset in~\Cref{ssec:coma}. 

\subsection{Constructing the space $\latentImm_{\bar{q}}$}\label{sec:basis}
To construct the bases of movements and body type deformations (expression and face identity for the COMA dataset, resp.) we interpret registered mesh sequences of motions (expressions, resp.) as paths in shape space whose tangent vectors are implicitly restricted to the space of valid motions.  We first collect meshes of the same pose (expression) from each identity and compute the (unrestricted) pairwise geodesics between these meshes with respect to our second-order Sobolev metric, where we use the Pytorch implementation of~\cite{hartman2023elastic}.
Note that these meshes show only moderate deformations and thus there are no difficulties with applying the unrestricted matching algorithm. We then collect the tangent vectors to these paths and perform PCA to define our basis of shape deformations. We could use a similar strategy for generating our basis of motion (expression) deformation, i.e., collect shapes with the same body type (face identity, resp.) and calculate the unrestricted pairwise geodesics between these meshes. This can, however, lead to unnatural motions for large movements. Instead, we take advantage of the available 4D data in our targeted application data sets. This allows us  to perform principle component analysis directly on the tangent vectors of those real data sequences to obtain a valid pose (expression) data basis. In the ablation study section~\ref{sec:ablation} we will compare the quality of the results for these two approaches, i.e., by using 4D data versus using only 3D data for the basis construction. We point out that a similar procedure has been used for the analysis of pre-registered human body motions in previous work by the authors~\cite{Pierson_2022_WACV}. We should also note that we here pre-construct the bases from a fixed predefined training set. Another possible approach, used for instance in \cite{muralikrishnan2023bliss} (albeit only for shape deformations), is to progressively enrich some initial estimation of a basis via a bootstrapping scheme, providing a possible alternative way to build shape/pose deformation bases from only a small training set of registered meshes.  

\subsection{Parameter selection} 
Next we describe the choice of parameters in our experiments. 
For the human bodies the coefficients for the $H^2$-metric were chosen to enforce close to isometric deformations that allow for some stretching and shearing to allow change in body type.  In the case of human body faces, we reduce the strecthing and shearing penalization, and enforce normal consistency. We added a small coefficient to the remaining terms to further regularize the deformations. The final six parameters for the $H^2$-metric are set to $(1, 1000, 100, 1, 1, 1)$ for human bodies and $(1, 10, 10, 10, 1, 1)$ for human faces. The basis size for both applications is as follows (the number of elements was chosen experimentally, cf. Section~\ref{sec:ablation}): the motion basis has $n=130$ elements, whereas the basis for the body type variation has only  $m=40$ elements. Furthermore, we perform sequential minimizations where the parameter $\sigma$ of the varifold term is decreased from $.4$ to $.025$ and the balancing term $\lambda$ is increased from $10^{2}$ to $10^{8}$. In the case of human faces, we needed only two minimizations with the parameter $\sigma$ of the varifold term at $.01$ and $.005$ and the balancing term $\lambda$ at $10^{6}$ and $10^{10}$.
 
\subsection{Evaluation methods}\label{sec:evaluationmethods}
In our experiments, we will evaluate the quality of the results using different similarity measures (distances) between the outputs of the different methods and the original scan. The ``shape'' matching is evaluated by comparing each method against the original scans using three different remeshing invariant similarity measures.  First, we evaluate the methods using the varifold metric introduced before. As our method minimizes this distance during the registration process, we include two additional metrics to avoid bias: the widely used  Hausdorff distance, which provides a good insight for the quality of a mesh reconstruction, but can be sensitive to single outliers present in low-quality scans and the Chamfer distance~\cite{fan2017point, groueix20183d}, which is more robust to such outliers.  

In our first experiment -- latent code retrieval -- we will in addition evaluate the quality of the obtained point correspondences -- in this section, we use data with given ground truth point correspondences. Therefore we will compute the mean squared error of each method to the ground truth registrations of the testing set. Unfortunately, one method (LIMP) does not return the same mesh structure as the ground truth registrations and thus we could not compare it this way. We thus add the geodesic error metric, which is equal to the mean of geodesic distances between estimated correspondences and their ground truth corresponding points on target meshes.  For a detailed description of all these evaluation metrics, we refer to the supplementary material.

\subsection{Comparison methods}
Finally, we will briefly describe the other state-of-the-art methods that we considered for comparison. A more detailed description of these methods can be found in the supplementary material.  We primarily compare to methods that rely on latent space learning for registration, interpolation, and extrapolation tasks and do not consider other methods that can potentially tackle the same tasks but without a low dimensional latent space~\cite{eisenberger2021neuromorph}, or that are specifically designed for other tasks~\cite{GLASSCvpr2022}. We compare our approach to LIMP~\cite{cosmo2020limp}, which models shape deformations using a variational auto-encoder with geodesic constraints; ARAPReg~\cite{huang2021arapreg}, which models deformations using an auto-decoder with regularization through the as rigid as possible energy; and 3D-Coded~\cite{groueix2018b}, which is similar to LIMP but with lighter training and without geometric loss regularization. LIMP and 3D-Coded both utilize a PointNet architecture as an encoder, which enables invariance to parameterization. On the other hand, ARAPReg recovers latent vectors within a registered setting utilizing the $L^2$ metric, which assumes that the target meshes possess an identical mesh structure as the model's output. To make this framework viable for our application we replace the $L^2$-metric by the varifold distance thereby extending ARAPReg to unregistered point clouds. We trained all three networks on the DFAUST dataset using reported training details from the respective papers. As a final comparison, we consider the FARM method~\cite{ShrecMelzi2019} from the regime of functional maps-based methods. Unfortunately, this method does not compute any interpolation or extrapolation of shape changes so we exclusively compare to this method for shape registration tasks.

\section{Experimental Results}\label{sec:HBResults}
In this section, we will demonstrate the capabilities of our framework in several different experiments.  For the human body scans, which will be our main targeted application, we will present a thorough comparison to several other state-of-the-art algorithms. Therefore we will provide quantitative and qualitative analysis of
the registration and point correspondence accuracy, the shape reconstruction quality, and the accuracy of interpolations and extrapolations to recreate real sequences of human motions. 
Furthermore, we give qualitative examples of our framework applied to random shape generation and motion transfer tasks. At the end of the section, we will present similar experiments for the COMA dataset, which consists of human face scans. The computational cost of our method is discussed in the supplementary material. 
\subsection{Mesh invariant latent code retrieval}
\begin{figure}
    \centering
    \small
    \begin{tabular}{|cm{6cm}|}\hline
        \begin{tabular}{l}BaRe-ESA \end{tabular}&\includegraphics[width=\linewidth]{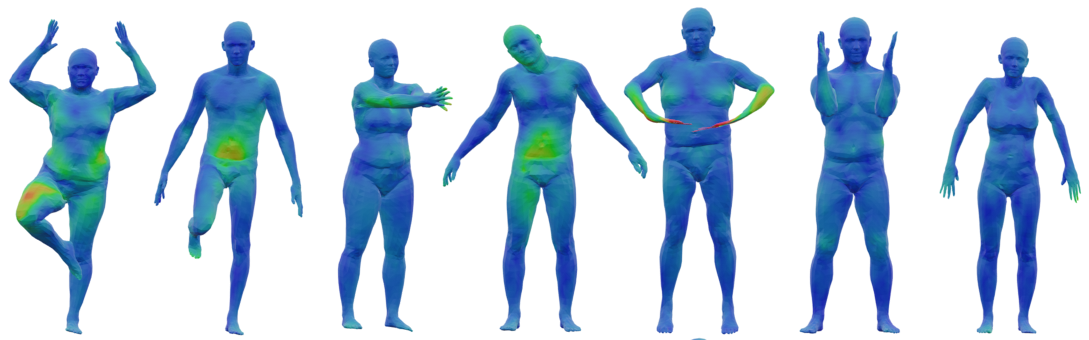}  \\\hline        
        \begin{tabular}{l}ARAPReg \end{tabular}&\includegraphics[width=\linewidth]{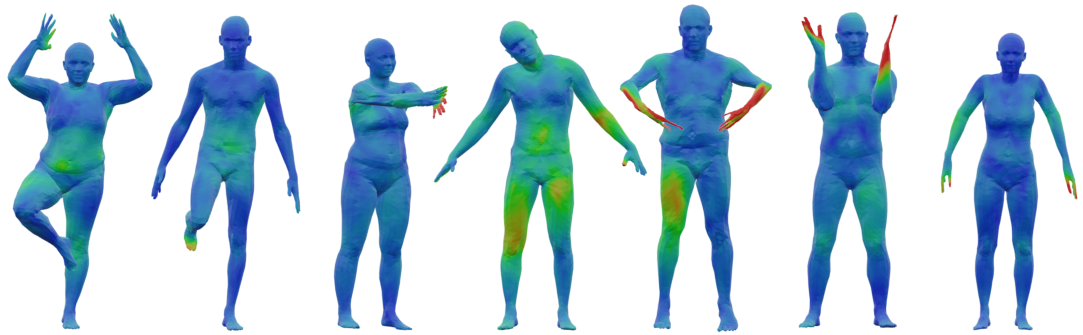} \\\hline
        \begin{tabular}{l}3D-Coded \end{tabular}&\includegraphics[width=\linewidth]{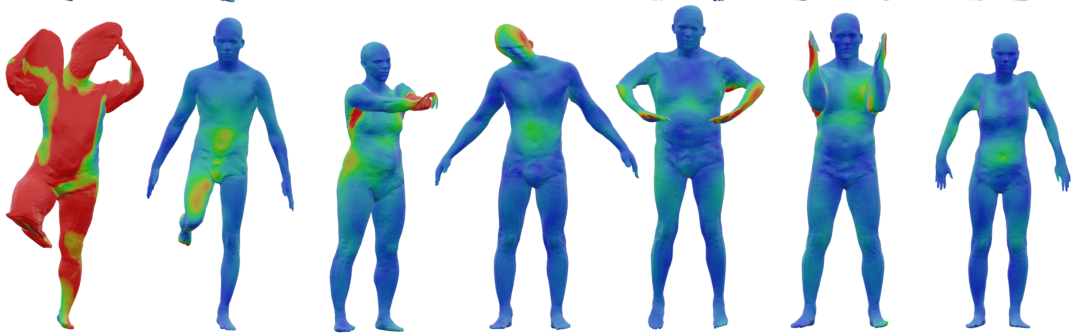} \\\hline
        \begin{tabular}{l}FARM \end{tabular}&\includegraphics[width=\linewidth]{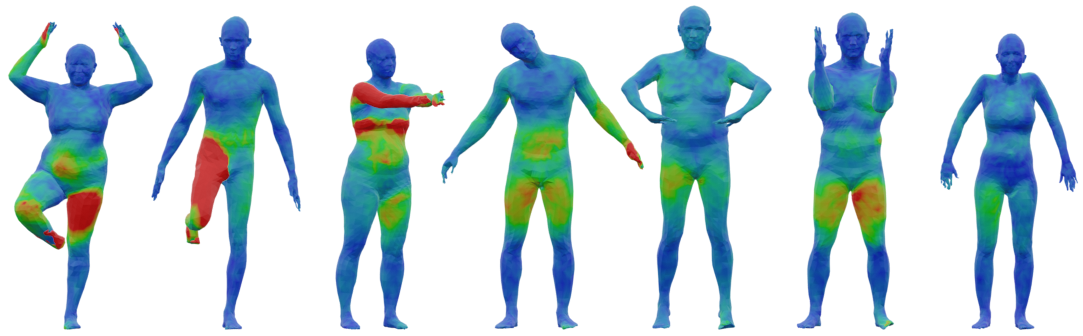} \\\hline
        \begin{tabular}{l}TRUTH \end{tabular}&\includegraphics[width=\linewidth]{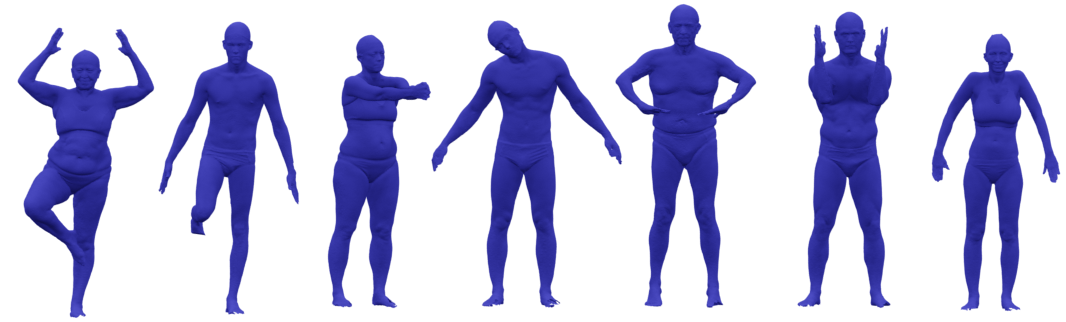} \\\hline
    \end{tabular}
    \caption{Registration of seven elements of FAUST using four methods trained on DFAUST. The registrations produced by 3D-Coded, FARM, and ARAPReg have regions with large deformation errors.  BaRe-ESA consistently produces a decent representation in all examples. The coloring of each mesh encodes the pointwise registration error from the ground truth with blue encoding 0mm error and red encoding $\geq15$mm error.  }
    \label{fig:registration}
\end{figure}

To demonstrate the effectiveness of our  latent code retrieval algorithm, cf. Section~\ref{sec:latentcodes}, we tested its performance on the three human body testing data sets as described in Section~\ref{sec:datasets}. 
In this experiment, we construct latent code representations with BaRe-ESA, LIMP, 3D-Coded, ARAPreg, and FARM and measure the distance from the reconstructed meshes to the original scans using the evaluation methods outlined in Section~\ref{sec:evaluationmethods}. In ~\Cref{fig:registration} we present a qualitative comparison of the obtained results. A quantitative comparison of the performance of the different methods is presented, with shape registration evaluation in~\Cref{tab:registration} and geometric reconstruction of the human shape in~\Cref{tab:reconstruction}. Both evaluations demonstrate that BaRe-ESA significantly outperforms the mesh autoencoder methods with respect to the registration and reconstruction evaluation metrics.

\begin{table}[t]
        \centering
        \scriptsize
        \setlength\tabcolsep{2pt} 
    \renewcommand{\arraystretch}{1.0}
    \begin{tabular}{l|ccccc|}
    &LIMP&ARAPReg&3D-Coded&FARM&BaRe-ESA\\ \hline
    MSE&NA&0.035&0.053&0.043&\textbf{0.014}\\
    Geodesic Error&0.15&0.031&0.038&0.038&\textbf{0.013}\\
    \end{tabular}
    \caption{Human body shape registration results. We compute the registration error on the FaustE data set. Where applicable, we compute the mean squared error (MSE) and geodesic error between each method's outputs and the ground truth registrations of FaustE.
    \label{tab:registration}}
\end{table}

\begin{table}[t]
        \centering
        \scriptsize
        \setlength\tabcolsep{2pt} % default value: 6pt
        \begin{tabular}{c}
        \begin{adjustbox}{max width=\textwidth}
        \aboverulesep=0ex
        \belowrulesep=0ex
        \renewcommand{\arraystretch}{1.0}
        \begin{tabular}{l|cc|cc|cc|}
        &\multicolumn{2}{c}{Hausdorff} &
        \multicolumn{2}{c}{Chamfer} &
        \multicolumn{2}{c|}{Varifold} \\
        & FAUST & SHREC & FAUST & SHREC & FAUST & SHREC\\\cmidrule{2-7}
        LIMP
        &0.23&0.17
        &0.098&0.070
        &0.073&0.057\\          
        ARAPReg
        &0.11&0.11
        &0.117&0.028
        &0.021&0.036\\     
        3D-Coded
        &\textbf{0.07}&\textbf{0.07}
        &0.020&\textbf{0.022}
        &0.023&\textbf{0.034}\\     
        BaRe-ESA
        &0.08&0.13
        &\textbf{0.019}&0.029
        &\textbf{0.014}&\textbf{0.034}\\     
        \end{tabular}
        \end{adjustbox}
    \end{tabular}
    
    \caption{Human body shape reconstruction results. We compute the Hausdorff, Chamfer, and Varifold reconstruction errors between the outputs of the methods and the original scans. We evaluate these methods on the FaustE and ShrecE testing sets.}
    \label{tab:reconstruction}
\end{table}

\subsection{Interpolation and Extrapolation Results}\label{sec:inter_extra}
\begin{figure}
    \centering
    \small
    \begin{tabular}{cm{5.75cm}}
        \begin{tabular}{l}BaRe-ESA \end{tabular}&\includegraphics[width=\linewidth]{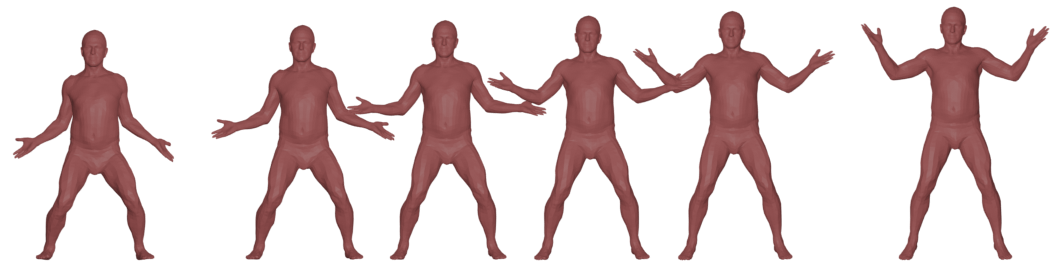}  \\\hline
        \begin{tabular}{l}3D-Coded \end{tabular}&\includegraphics[width=\linewidth]{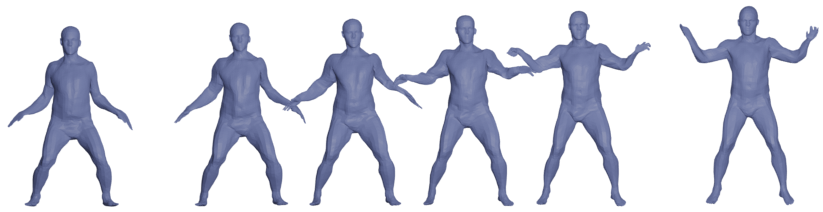}  \\\hline
        \begin{tabular}{l}LIMP \end{tabular}&\includegraphics[width=\linewidth]{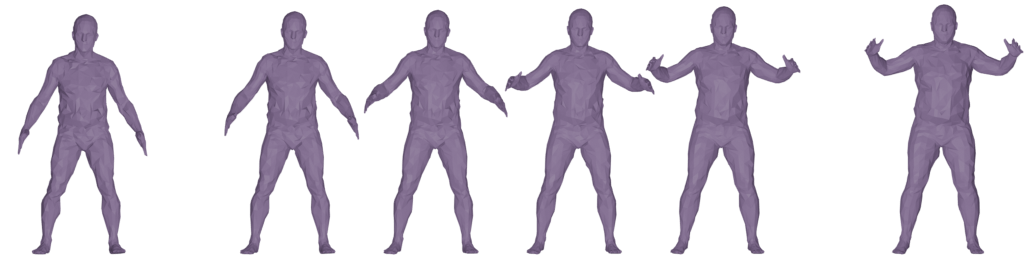}  \\\hline
        \begin{tabular}{l}ARAPreg \end{tabular}&\includegraphics[width=\linewidth]{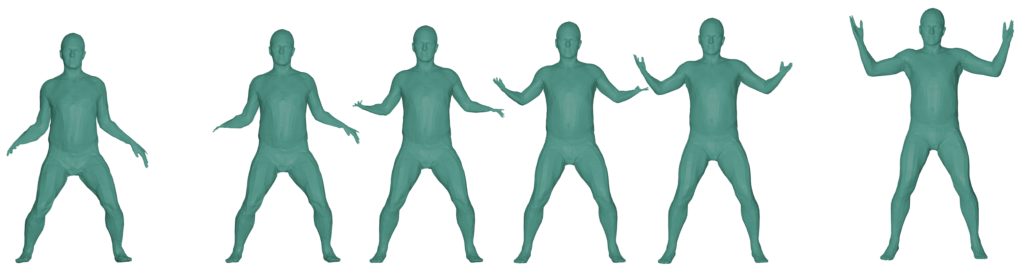} \\\hline
        \begin{tabular}{l}GROUND\\TRUTH\end{tabular}& \includegraphics[width=\linewidth]{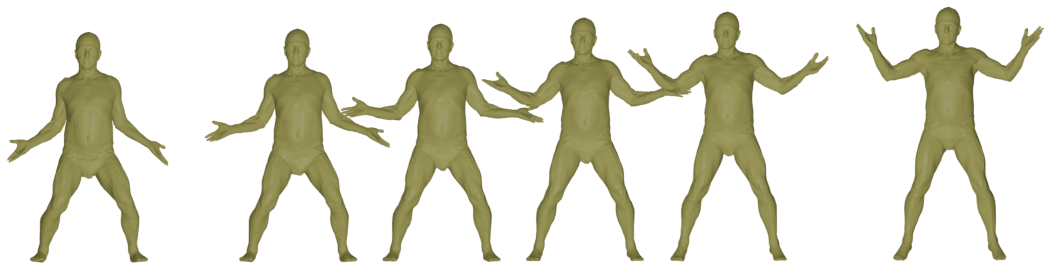}  \\
        &$\,\,\underbracket{\hspace*{.75cm}}_{\text{Source}}\,\quad\underbracket{\hspace*{3cm}}_{\text{Interpolation}}\,\quad\underbracket{\hspace*{.75cm}}_{\text{Target}}$
    \end{tabular}
    \caption{Interpolation results comparison between our method, LIMP, ARAPReg and the Ground Truth from DFAUST. While the path produced by LIMP does not properly register the endpoints and the path produced by ARAPreg does not stay in the space of human bodies, BaRe-ESA successfully produces a path of human shapes whose endpoints match the source and target shapes.  }
    \label{fig:interpolation_HB}
\end{figure}

Next, we turn our attention to the interpolation problem for human bodies, i.e., the task of constructing a deformation between two different human body poses, that follows a ``realistic'' motion pattern. We use the start and end points of our 10 test mini-sequences from the DFAUST data set as the input for these experiments. This allows us to compare the obtained results to the full mini-sequences, seen as a ground truth motion (see the supplementary material for their corresponding animations). In ~\Cref{fig:interpolation_HB}, we show a qualitative comparison of our method to ARAPReg, 3D-Coded, and LIMP. Our method is successful at recovering the latent codes that represent the endpoints and producing interpolations that remain in the space of human shapes. We further perform a quantitative comparison of the methods by measuring the distance to the ground-truth sequences at each break point with respect to the evaluation metrics given in Section \ref{sec:evaluationmethods}; these results are displayed in~\Cref{tab:interpolation_HB}. One can clearly observe that our method again outperforms the other methods both qualitatively and quantitatively.

\begin{table*}[t]
    \centering
    \scriptsize
    \setlength\tabcolsep{2pt} % default value: 6pt
    \begin{tabular}{c}
        \begin{adjustbox}{max width=\textwidth}
            \aboverulesep=0ex
            \belowrulesep=0ex
            \renewcommand{\arraystretch}{1.0}
            \begin{tabular}[t]{c|cccc|cccc|cccc}
            \multicolumn{1}{c}{} & 
            \multicolumn{12}{c}{Interpolation}  \\
            \cmidrule{2-13}
            \multicolumn{1}{c}{} &
            \multicolumn{4}{c}{Hausdorff} &
            \multicolumn{4}{c}{Chamfer} &
            \multicolumn{4}{c}{Varifold}  \\
            & LIMP & ARAPReg & 3D-Coded & BaRe-ESA  & LIMP & ARAPReg & 3D-Coded & BaRe-ESA & LIMP & ARAPReg & 3D-Coded & BaRe-ESA \\
                \midrule
                punching &
                4.650 & 4.786 & 4.882 & \textbf{1.009} &
                1.488 & 1.553 & 1.694 & \textbf{0.350} &
                1.373 & 0.869 & 1.182 & \textbf{0.252}  \\
                 
                running on spot &
                2.045 & 0.977 & 1.357 & \textbf{0.820} &
                1.026 & \textbf{0.334} & 0.454 & 0.475 &
                0.786 & \textbf{0.359} & 0.441 & 0.372 \\
                
                running on spot b &
                2.367 & 1.726 & 1.931 & \textbf{1.134} &
                1.039 & 0.653 & 0.706 & \textbf{0.548} &
                0.767 & 0.488 & 0.545 & \textbf{0.366} \\
                 
                shake arms &
                1.698 & 1.145 & 1.456 & \textbf{0.847} &
                0.764 & 0.327 & 0.496 & \textbf{0.326} &
                0.672 & 0.206 & 0.391 & \textbf{0.180} \\
                 
                chicken wings &
                4.774 & 4.926 & 4.951 & \textbf{1.289} &
                2.058 & 2.356 & 2.535 & \textbf{0.636} &
                1.276 & 0.666 & 0.807 & \textbf{0.296} \\
                 
                knees &
                12.898 & 2.797 & 19.593 & \textbf{0.718} &
                8.803 & 0.496 & 18.153 & \textbf{0.461} &
                2.067 & 0.627 & 1.925 & \textbf{0.338} \\
                 
                knees b &
                5.516 & 1.055 & \textbf{0.738} & 1.995 &
                1.862 & 0.262 & \textbf{0.249} & 0.693 &
                0.748 & 0.298 & \textbf{0.279} & 0.347 \\
                 
                jumping jacks &
                1.397 & 1.320 & 1.164 & \textbf{0.811} &
                0.762 & 0.350 & 0.380 & \textbf{0.333} &
                0.769 & 0.253 & 0.329 & \textbf{0.229} \\
                 
                jumping jacks b &
                3.518 & 2.140 & 2.607 & \textbf{1.482} &
                1.635 & \textbf{0.672} & 1.005 & 0.692 &
                0.882 & 0.369 & 0.523 & \textbf{0.254}  \\
                 
                one leg jump &
                1.931 & 0.748 & 0.853 & \textbf{0.616} &
                0.806 & 0.274 & 0.281 & \textbf{0.221} &
                0.739 & 0.329 & 0.367 & \textbf{0.264} \\
                \midrule
                mean &
                4.079 & 2.162 & 3.953 & \textbf{1.072} &
                2.024 & 0.728 & 2.595 & \textbf{0.474} &
                1.008 & 0.447 & 0.679 & \textbf{0.290} 
        \end{tabular} 
        \end{adjustbox}
      
    \end{tabular}
    \vspace{-3pt}
    \caption{Full interpolation comparison on 10 DFAUST sequences. The Hausdorff, Chamfer and varifold distance are computed against ground truth sequences.\vspace{10pt}}
    \label{tab:interpolation_HB}
    \vspace{-18pt}
\end{table*}

\begin{figure}
    \small
    \begin{tabular}{cm{5.75cm}}
        \begin{tabular}{l}BaRe-ESA \end{tabular}&\includegraphics[width=\linewidth]{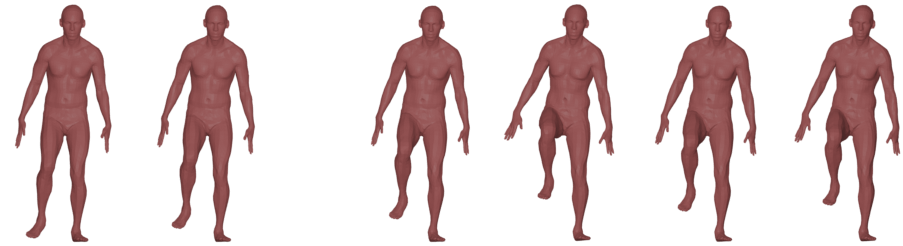}  \\\hline
        \begin{tabular}{l}3D-Coded \end{tabular}&\includegraphics[width=\linewidth]{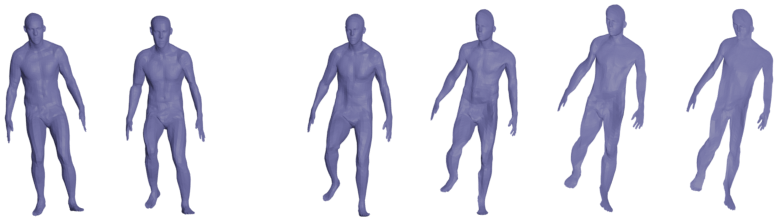}  \\\hline
        \begin{tabular}{l}LIMP \end{tabular}&\includegraphics[width=\linewidth]{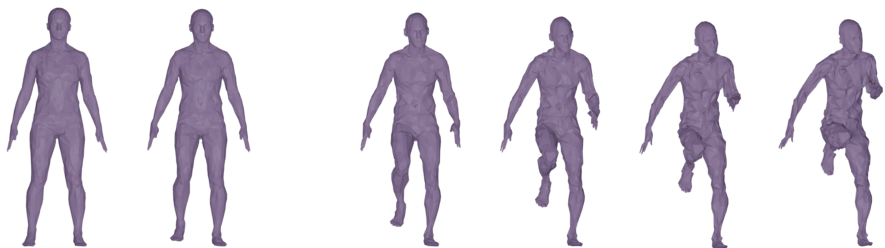}  \\\hline
        \begin{tabular}{l}ARAPreg \end{tabular}&\includegraphics[width=\linewidth]{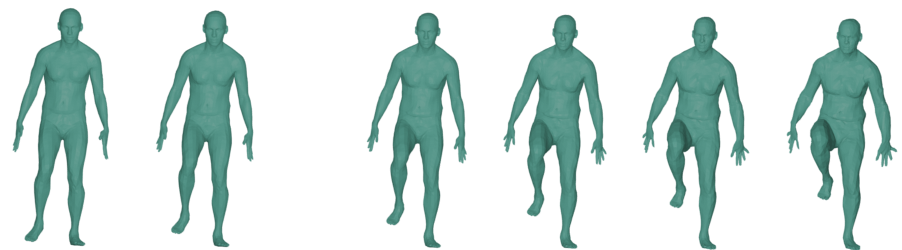} \\\hline
        \begin{tabular}{l}GROUND\\TRUTH\end{tabular}& \includegraphics[width=\linewidth]{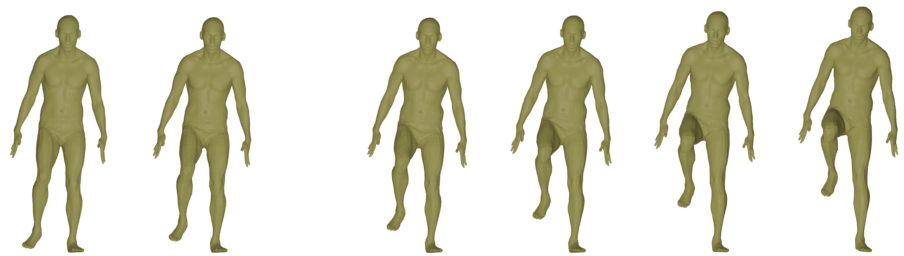}  \\
        &$\underbracket{\hspace*{1.75cm}}_{\text{Input }}\qquad\underbracket{\hspace*{3.5cm}}_{\text{Extrapolation}}$
    \end{tabular}
    \caption{Extrapolation results comparison between our method, LIMP, ARAPReg and DFAUST Ground Truth. While all methods capture the primary motion of lifting a leg, the extrapolations of LIMP and ARAPreg include extraneous motions of arms and slight changes in body type.}
    \label{fig:extrapolation_HB}
\end{figure}

Next, we consider the related problem of human body shape extrapolation, i.e., the task of predicting the future movement given a body shape and an initial movement (deformation). We consider again the 10 mini-sequences from the DFAUST dataset. We then recover the latent codes of the first two meshes in the sequence and  use the first latent code and the difference of the codes as input to our method. In \Cref{fig:extrapolation_HB}, we present again a qualitative comparison of our results to the extrapolations computed using LIMP, 3D-Coded, and ARAPreg (see the supplementary material for their corresponding animations). One can see that our method is successful at producing extrapolations that capture the correct motion of the mesh without any extraneous motions that stay in the space of human bodies. As with the interpolation comparison, we measure the distance to the ground-truth sequences at each breakpoint and display the results of the quantitative comparison in~\Cref{tab:extrapolation_HB}. Similar to the previous experiments, our method significantly outperforms the other methods.

\begin{table*}[t]
    \centering
    \scriptsize
    \setlength\tabcolsep{2pt} % default value: 6pt
    \begin{tabular}{c}
        \begin{adjustbox}{max width=\textwidth}
            \aboverulesep=0ex
            \belowrulesep=0ex
            \renewcommand{\arraystretch}{1.0}
            \begin{tabular}[t]{c|cccc|cccc|cccc}
            \multicolumn{1}{c}{} & 
            \multicolumn{12}{c}{Extrapolation} \\
            \cmidrule{2-13}
            \multicolumn{1}{c}{} &
            \multicolumn{4}{c}{Hausdorff} &
            \multicolumn{4}{c}{Chamfer} &
            \multicolumn{4}{c}{Varifold} \\
            & LIMP & ARAPReg & 3D-Coded & BaRe-ESA  & LIMP & ARAPReg & 3D-Coded & BaRe-ESA & LIMP & ARAPReg & 3D-Coded & BaRe-ESA \\
                \midrule
            punching &                     
4.232 & 8.142 & 5.792 & \textbf{4.952} &
1.494 & 2.436 & 2.685 & \textbf{1.424} &
1.506 & 1.551 & 1.441 & \textbf{0.901}\\                     
            running on spot &                     
2.846 & 3.437 & 2.340 & \textbf{1.973} &
1.184 & 1.617 & 1.095 & \textbf{1.071} &
0.805 & 1.135 &\textbf{ 0.607} & 0.788  \\                  
            running on spot b &                     
2.404 & 2.435 & 1.699 & \textbf{1.392} &
1.122 & 0.828 & \textbf{0.759} & 1.073 &
0.787 & 0.749 & \textbf{0.515} & 0.839 \\  
            shake arms &                     
2.090 & 2.737 & 1.734 &\textbf{1.109} &
1.017 & 0.892 & 0.630 &\textbf{0.421} &
0.771 & 0.528 & 0.520 & \textbf{0.330} \\                  
            chicken wings &                     
4.778 & 12.790 & 5.224 & \textbf{4.952} &
2.230 & 5.127 & 2.536 & \textbf{2.373} &
1.475 & 1.673 & \textbf{1.117} & 1.121 \\                  
            knees &                     
42.529 & 6.713 & 49.820 & \textbf{3.632} &
32.943 & \textbf{1.144} & 39.805 & 2.074 &
6.794 & 1.470 & 2.699 & \textbf{1.428}  \\                  
            knees b &                     
9.993 & 2.418 & \textbf{1.942} & 3.455 &
3.343 & \textbf{0.554 }& 1.050& 1.323 &
1.380 & 0.633 & \textbf{0.506} & 0.722 \\  
            jumping jacks &                     
4.116 & 5.873 & 8.696 & \textbf{2.149} &
1.767 & 2.345 & 6.449 & \textbf{0.917} &
1.099 & 1.038 & 0.699 &\textbf{0.476} \\  
            jumping jacks b &                     
2.219 & 3.519 & 1.759 & \textbf{1.436} &
0.992 & 0.984 & 0.702 & \textbf{0.411} &
0.765 & 0.623 & 0.498 & \textbf{0.270} \\  
            one leg jump &                     
2.195 & 1.970 & 1.989 & \textbf{0.867} &
0.906 & 0.757 & 0.915 & \textbf{0.427} &
0.758 & 0.858 & 1.800 & \textbf{0.540} \\  
            \midrule
            mean &                     
7.740 & 5.004 & 8.100 & \textbf{2.592} &
4.700 & 1.668 & 5.663 & \textbf{1.151} &
1.614 & 1.026 & 1.040 & \textbf{0.742 }\\  
        \end{tabular} 
        \end{adjustbox}
      
    \end{tabular}
    \vspace{-3pt}
    \caption{Full extrapolation comparison on 10 DFAUST sequences. The Hausdorff, Chamfer and varifold distance are computed against ground truth sequences.\vspace{10pt}}
    \label{tab:extrapolation_HB}
    \vspace{-18pt}
\end{table*}

\subsection{Motion Transfer and Random Shape Generation}
\begin{figure}
\centering
\begin{tabular}{m{1.2cm}|m{7cm}}
    \includegraphics[width=\linewidth]{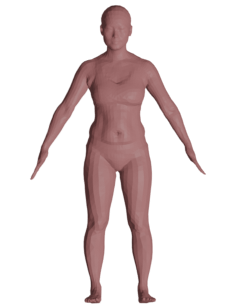}&\includegraphics[width=\linewidth]{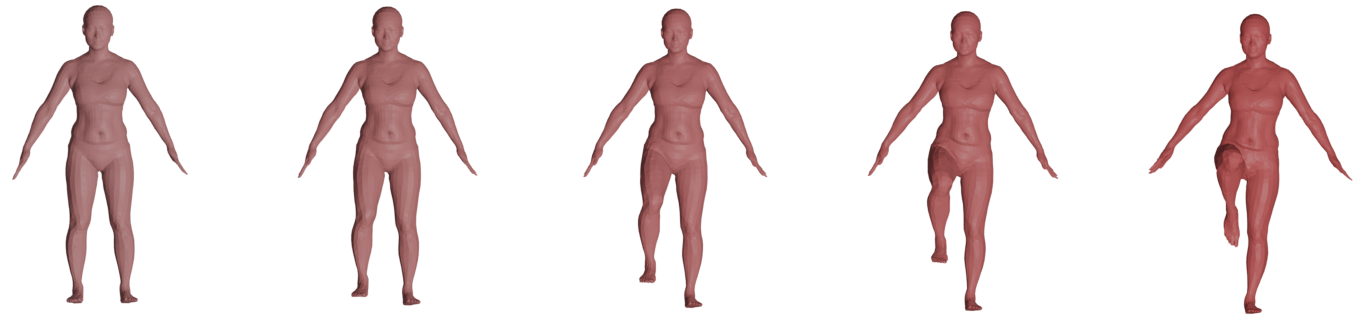}  \\ 
    \includegraphics[width=\linewidth]{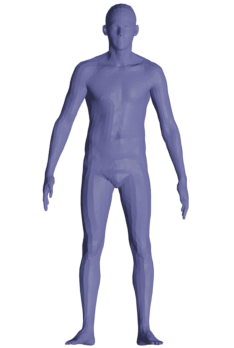}&\includegraphics[width=\linewidth]{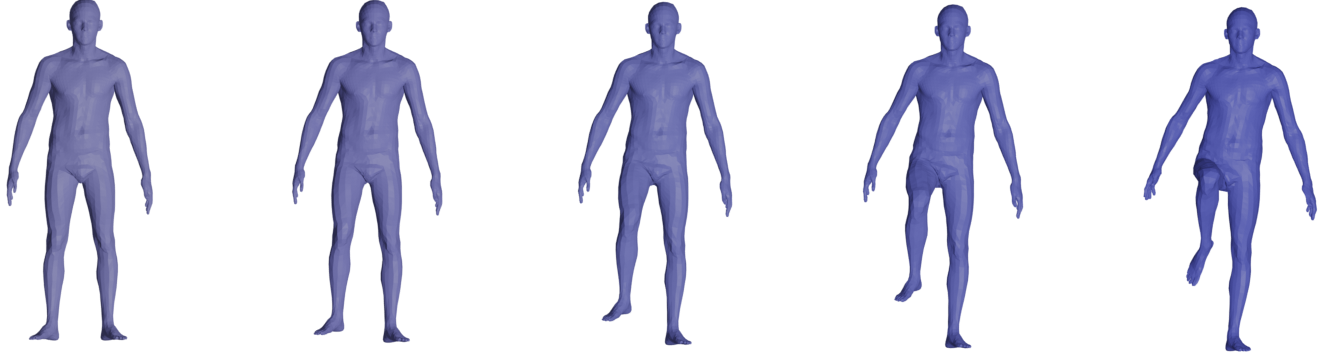}  \\ 
    \includegraphics[width=\linewidth]{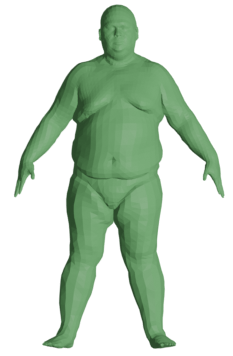}&\includegraphics[width=\linewidth]{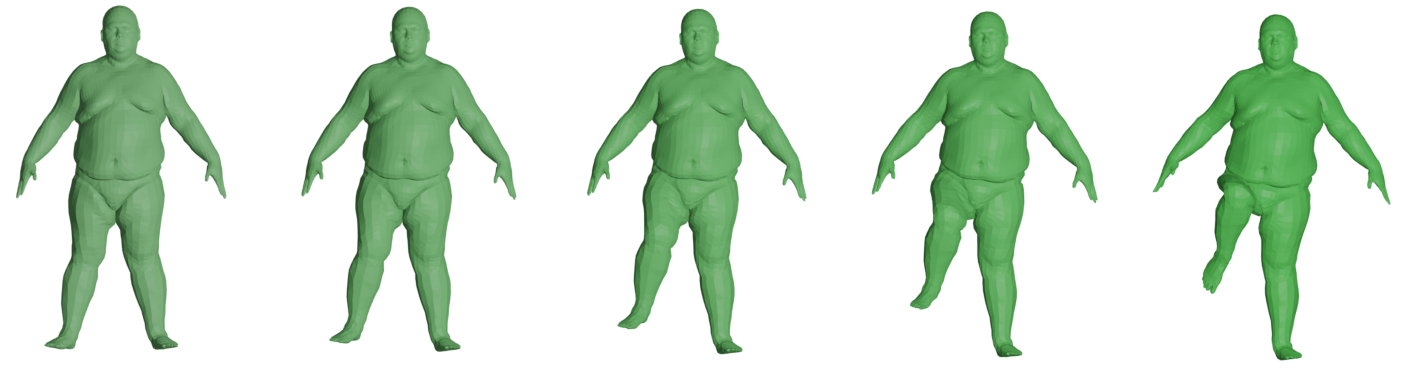}  \\
\end{tabular}
\caption{Motion Transfer: We display the original motion in the top row and the transfer of the motion to the target shapes in the second and third row.}
\label{fig:motion_transfer_HB}
\end{figure}

\begin{figure}
\centering
\includegraphics[width=\linewidth]{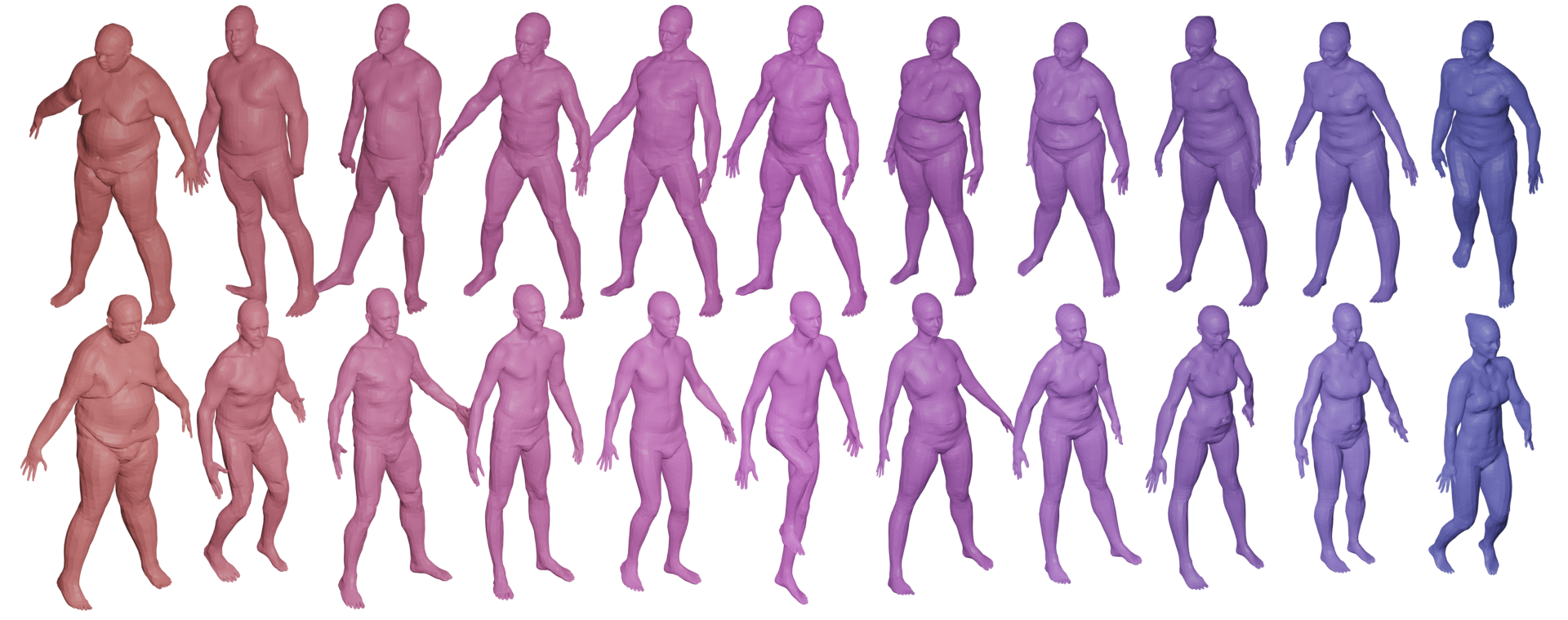}
\caption{Random Shapes: 22 random shapes generated using a Gaussian mixture model on the space of initial velocities.}
\label{fig:random_HB}
\end{figure}

As two further examples of the capabilities of our framework, we present applications to motion transfer and random shape generation. For the motion transfer, we first represent a motion as a sequence of latent codes and then we simply replace the shape coefficients of each element of the sequence with the shape coefficients of the target shape. An example of this method in action is displayed in \Cref{fig:motion_transfer_HB}.
Another possible application of our framework is random shape generation. The idea is to use a data-driven distribution on the human shape tangent space. 
Therefore we first perform latent code retrieval on a subset of DFAUST. We then compute, 
the initial tangent vector of each of these paths in the latent space, separated in pose and shape components.
For each of these collections of tangent vectors, we fit a Gaussian mixture model, which is popular for generating human shapes~\cite{bogo2016keep, omran2018neural}. We used 10 and 6 components respectively, which proved to be sufficient to get visually satisfying random shapes. The generation process consists of sampling a pose and shape vector in the tangent space and solving the corresponding geodesic initial value problem from the template in the direction of the generated vector. We display a selection of 13 generated shapes in~\Cref{fig:random_HB}.
\subsection{Application to Human Faces}\label{ssec:coma}
\begin{figure}
    \centering
    \begin{tabular}{cm{5.75cm}}
    \includegraphics[width=0.9\linewidth]{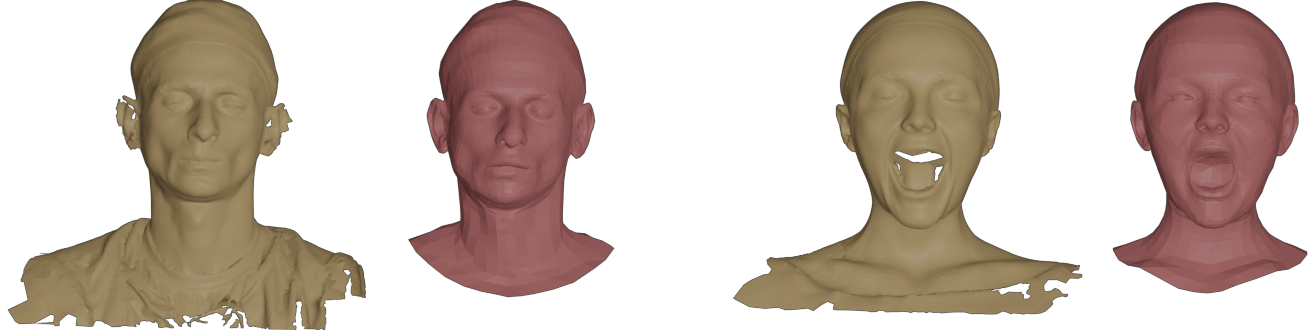}\\ \hline\\
    \includegraphics[width=0.9\linewidth]{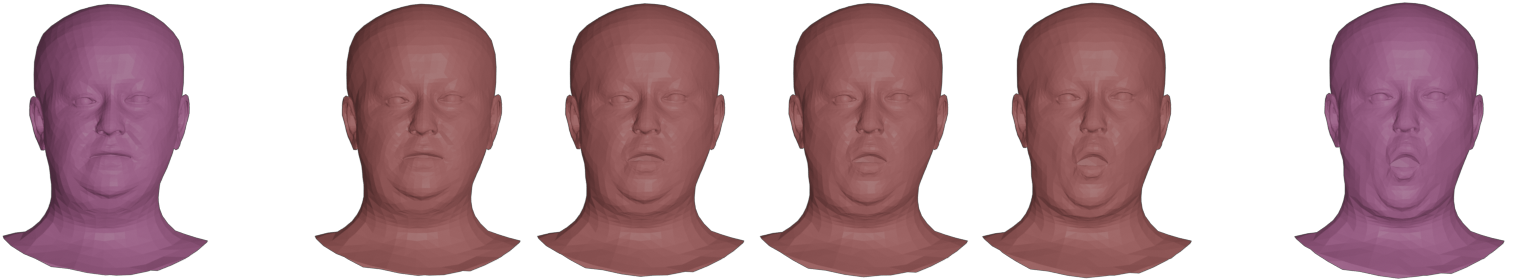}\\ \\\hline\\
    \includegraphics[width=0.9\linewidth]{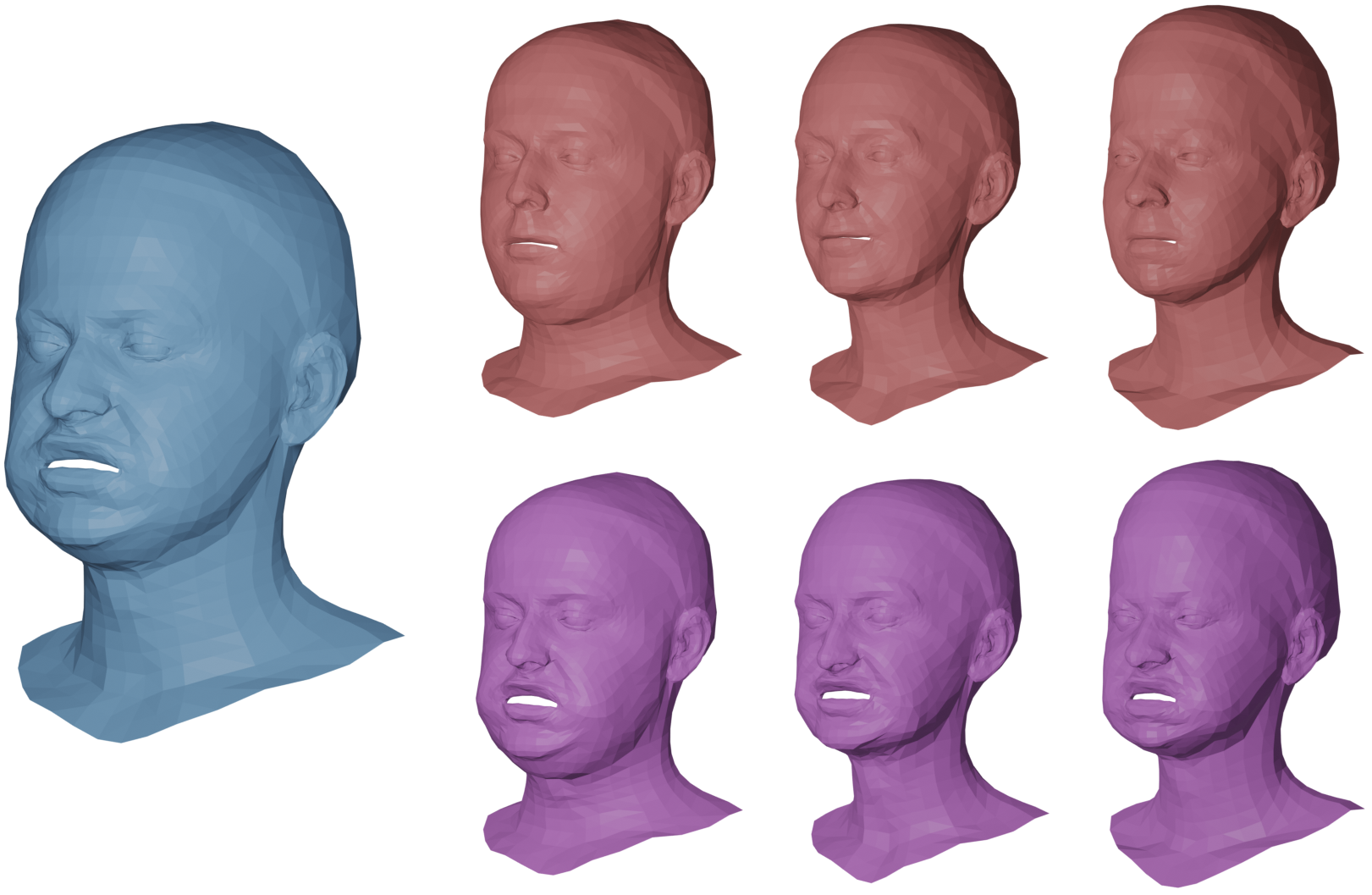}
    \end{tabular}
    \caption{Experimental results for COMA faces. Here we present several qualitative results from this framework applied to the COMA dataset. In the first row, we show our latent code reconstruction (red) of two different noisy scans of human faces (yellow). In the second row, we display an example of a solution to a geodesic boundary value problem to interpolate between two shapes (purple). In the third row, we display an example of expression transfer in our framework. The blue mesh on the left represents a target face registered with our framework, each red mesh on the right represents three additional identities and the purple meshes below represent the transfer of the expression onto these identities.}
    \label{fig:coma_rec}
\end{figure}

Finally in Figure~\ref{fig:coma_rec} we showcase the capabilities of our framework in the context of human face scan analysis from the COMA dataset. In this figure, we show two latent code reconstructions of two different noisy scans of human faces, an example of an interpolation between two different expressions and an expression transfer to a different identity. 
Additional examples of registration, interpolations, and a qualitative comparison to the other deep learning methods are shown in the supplementary material. We do not present a full quantitative comparison as we were not able to get satisfactory results with most of the other methods, with the exception being ARAPReg which performed comparably to our method. One reason for the better performance of ARAPReg as compared to LIMP and 3D-Coded is probably the use of the varifold distance in our adapted implementation of this approach, the original implementation of ARAPReg not being capable of dealing with unregistered data. The other learning-based methods (LIMP and 3D-coded) use instead the Chamfer distance. We believe that this might be one source of the significantly worse performance of these methods on the COMA dataset.

\section{Ablation Studies}\label{sec:ablation}
Within this section, we conduct a sequence of ablation studies to validate our selections regarding the number of shapes and pose basis elements, the shape-matching function, and the Riemannian metric employed in the calculation of path energy. Finally, we also consider the performance of BaRe-ESA using only static (3D) data to compute our basis.
\subsection{Choice of basis size}
Our first ablation study considers the choice of basis size: in Table~\ref{tab:ablation_basis}, we present the registration error corresponding to various numbers of shape and pose basis vectors. Each value in the table is determined by optimizing the latent vector reconstruction energy, as detailed in \ref{eq:latent_code3}, for an identical number of optimization iterations. The obtained results suggest that our choice of basis size provides the ideal balance of minimizing the latent space dimension while maximizing the expressibility of the obtained shape model. 
\begin{table}[htbp]
    \centering
    \scriptsize
    \begin{tabular}{c|cccc}
         &10&70&130&190  \\\hline
         10&0.090&0.072&0.016&0.015  \\ 
         40&0.092&0.068&\textbf{0.014}&0.015  \\ 
         70&0.089&0.063&0.015&0.014 \\ 
         100&0.087&0.062&0.016&0.016
    \end{tabular}
    \caption{Ablation study on the number of basis elements. We report the registration errors where we vary the number of pose and shape basis vectors of used in the matching process.The basis vectors are derived from training with DFAUST and the errors of the methods are calculated using data from FAUST. }
    \label{tab:ablation_basis}
\end{table}
\subsection{Choice of matching functional and latent space metric}
To justify our choice of shape matching functional and path energy, we compute the mean registration errors and interpolation errors for different combination of shape matching and path energy functionals. In particular, we experiment with replacing the varifold distance with the Chamfer distance and the elastic energy with the Euclidean distance on the latent space. The results of these experiments are reported in Table~\ref{tab:ablation_functions}. First, we compare the performance of the Chamfer and varifold distances and demonstrate that the choice of the varifold metric leads to significantly lower registration errors than the  Chamfer distance. In a second experiment, we demonstrate that the elastic matching energy produces significantly lower interpolation errors than that of an Euclidean path energy on the latent space. 
\begin{table}[htbp]
    \centering
    \scriptsize
    \begin{tabular}{ll||c|ccc}
         Shape Matching&Path&Registration&\multicolumn{3}{c}{Interpolation Errors} \\
         Term&Energy&Error&Haus.&Cham.&Var.\\\hline
         Chamfer&$H^2$&0.032&1.273&0.510&0.348 \\ 
         Varifold&Euclidean&0.015&1.787&0.632&0.335 \\
         Varifold&$H^2$&\textbf{0.014}&\textbf{1.072}&0.474&0.290 \\ 
    \end{tabular}
    \caption{Ablation study on the shape matching and path energy functions. We report the registration and interpolation errors for each combination of shape matching and path energy functions. For these experiments, we train the method using DFAUST and tested with FAUST. }
    \label{tab:ablation_functions}
\end{table}
\subsection{3D vs 4D training data}
In our final ablation study, we compare the results of our framework where we generate our motion basis (1) using real 4D data as done in the experiments of the previous section and (2) by computing elastic geodesics between scans of meshes of the same identity in different poses, i.e., only assuming the existence of 3D training data and generate the necessary motion and deformation paths with a standard elastic matching algorithm. In~\cref{tab:3dvs4d} we present the mean error for the interpolation problem for the 
same ten sequences from DFAUST as considered in~\cref{sec:inter_extra}: comparing these results with those of~\cref{tab:interpolation_HB} one can observe that the interpolation error is significantly worse as compared to the error obtained by training BaRe-ESA with 4D-data, but that we still outperform the three other baselines (LIMP, ARAPReg, 3D-Coded) in all three measures of performance (Hausdorff, Chamfer, varifold). We observed a similar behavior for the other tasks. 

\begin{table}[htbp]
    \centering
    \scriptsize
    \begin{tabular}{c|ccc}
         &Hausdorff&Chamfer&Varifold  \\\hline
         mean&2.004&0.683&0.405  
    \end{tabular}
    \caption{Interpolation results using only data from Faust (3D) for training. The mean errors are calculated for the same ten sequences from DFAUST as presented in ~\cref{sec:inter_extra}. }
    \label{tab:3dvs4d}
\end{table}

\section{Conclusions}\label{sect:conclusions}

In this paper, we proposed a general framework for basis restricted elastic shape analysis on the space of unregistered surfaces. We demonstrated superior performance compared to state-of-the-art methods in various tasks such as shape registration, interpolation, motion transfer, and random pose generation. Our framework utilizes a finite-dimensional latent space representation, which we equip with a non-Euclidean Riemannian metric inherited from the family of elastic metrics. This allows for a simplified representation of shape space while preserving the ability to compare surfaces modulo shape preserving transformations, i.e., our approach does not assume pre-registered surfaces or consistent mesh structures, making it applicable to a wide range of surface meshes with real data. Furthermore, the framework shows good generalization properties and does not require a substantial amount of training data.
The paper presents qualitative examples and quantitative analysis to support the effectiveness of the proposed framework in various experiments, including human body shape and pose data as well as human face scans. 

Lastly, we want to mention limitations and corresponding open directions for future work. Therefore we first point out that, as compared to some of the other latent space methods, the non-Euclidean nature of the latent space comes at the price of solving optimization problems to estimate interpolated or extrapolated geodesic paths, which can encumber to significant computational cost for large data applications. A possible way around this limitation would be to train neural networks in a supervised setting to learn the geometry of the latent space, i.e., to approximate the solutions of the interpolation and extrapolation problems.  

Finally, we want to mention a simple yet potentially relevant extension of our  model, namely to introduce  distinct Sobolev Riemannian metrics on the different shape modalities, e.g. for the shape change and the pose change deformation field in the human body motions. This comes with the idea of adapting the metric to the different nature of those deformations, and thus even better disentangling these quantities. 

\section{Acknowledgments}
This work was supported by ANR projects 16-IDEX-0004 (ULNE) and by ANR-19-CE23-0020 (Human4D); by NSF grants DMS-1912037, DMS-1953244, DMS-1945224 and DMS-1953267, and by FWF grant FWF-P 35813-N.

\clearpage

\section{Appendix}
\setcounter{section}{0}
\section{Geodesic distance bounds}
In this section we will study the induced geodesic distance of the second order Sobolev metric used in this article. For a finite dimensional Riemannian manifold the induced geodesic distance is always a true distance function, i.e., it is symmetric, satisfies the triangle inequality and is non-degenerate. This last property can, however, fail in infnite dimensions: there exists Riemannian geometries such that the geodesic distance between distinct points is zero or it might even vanish on the whole manifold. This startling phenomenon was first observed by Eliashberg and Polterovich~\cite{eliashberg1993bi} for the $H^{-1}$ metric on the symplectomorphism group and later by Michor and Mumford for the $L^2$ metric on spaces of immersions and diffeomorphisms~\cite{michor2005vanishing,bauer2012vanishing}. In the following theorem, we will prove that, under certain conditions on the parameters, the geodesic distance of the family of elastic Riemannian metric used in this article is non-degenerate:
\begin{theorem}
\emph{Let $a_0>0$ and let either $a_1,b_1,c_1,d_1>0$ or $a_2>0$ then the induced geodesic distance of the metric $G$ on the space $\Imm$ is non-degenerate, i.e., for any two surfaces $q_0,q_1\in \Imm$ with $q_0\neq q_1$ we have}
$
d_G(q_0,q_1)>0
$.\end{theorem}
%For the space space $\mathcal S$ this result is known, see eg.~\cite{bauer2011sobolev}; for the space of immersions it is, to the best of our knowledge, new:  
\newline
\begin{proof}
We start with the case that $a_1, b_1,c_1,d_1>0$. For this case we will make use of a generalization of the SRNF~\cite{JermynECCV2012,jermyn2017elastic} as introduced in~\cite{Su_2020_CVPRW}. To be more specific in~\cite{Su_2020_CVPRW} they considered the mapping 
\begin{equation}
\begin{aligned}
	\mathcal{Q}: \Imm&\to \operatorname{Met}(\pspace)\times C^{\infty}(\pspace, \R^3)\\
	q&\mapsto (q^*\langle.,.\rangle,\psi_q),
 \end{aligned}
\end{equation}
where $\operatorname{Met}(\pspace)$ denotes the space of all Riemannian metrics on $\pspace$ and where $\psi_q$ denotes the SRNF of $q$. On the space of all Riemannian metrics there exists a one parameter family of Riemannian metrics $G^E$, called the Ebin or DeWitt metric~\cite{deWitt67,ebin1970manifold}. Among other beneficial properties this Riemannian metric admits an explicit formula for its corresponding geodesic distance as derived by Clarke~\cite{clarke2010metric} and Michor-Meldrano~\cite{gil1991riemannian}. For the precise formula we refer to~\cite[Theorem 2]{Su_2020_CVPRW}. For the purpose of this proof it is only important that this distance is non-degenerate, i.e., $d_{G^E}(g_0,g_1)>0$ if $g_0\neq g_1$. On the second factor of the image of $Q$, i.e., on $C^{\infty}(\pspace,\R^3)$ we consider the standard non-invariant $L^2$ inner product as a Riemannina metric. This has again an explicit expression for the geodesic distance given by $d_{L^2}(\psi_1,\psi_2)=\|\psi_1-\psi_2\|^2_{L^2}$. The relevance of these results for our family of metrics can be found in the fact, that the pull-back of this product Riemannian metric via the mapping $Q$ yields exactly the Riemannian metric $G$ with parameters $a_0=d_1=a_2=0$ and $a_1,b_1,c_1\neq 0$ (depending on the parameter choice in the DeWitt metric and of the weighting of the two Riemannian metrics on the product space, see~\cite[Theorem 3]{Su_2020_CVPRW} for the precise statement of this result). 

Unfortunately the image of the map $Q$ in the product space $\operatorname{Met}(\pspace)\times C^{\infty}(\pspace, \R^3)$ is far from being totally geodesic and thus we cannot directly calculate the geodesic distance of the metric $G$ via this transform. Nevertheless, this construction still provides a lower bound for the geodesic distance of $G$ on $\Imm$, i.e., we have:
\begin{equation}
d_G(q_0,q_1)\geq d_{G^E}(g_0,g_1)+\|\psi_0-\psi_1\|^2_{L^2},
\end{equation}
where $(g_i,\psi_i)=Q(q)$. Next we note, that $Q(q_0)=Q(q_1)$ if and only if $q_0$ and $q_1$ differs only by a translation and we have shown that the geodesic distance of the elastic metric $G$ is non-degenerate on the quotient space $\Imm/{\operatorname{translation}}$. It remains to deal with the case that $q_0=q_1+v$ for some $v\in \mathbb R^3$. In this case the immersions $q_0$ and $q_1$ are also different elements in the quotient shape space $\mathcal S$ of unparametrized immersions, where the non-degeneracy has been shown using an area-swept-out-bound, see~\cite{bauer2011sobolev}. This concludes the proof assuming $a_1,b_1,c_1,d_1>0$. It remains to prove the result under the assumption that $a_2>0$, but in this situation the result follows directly from the above and the Sobolev embedding theorem.   
\end{proof}

\section{Discretization of invariant $H^2$ metrics}
In this section, we detail the computation of the Riemannian metric term $G_q(h,h)$ for discrete meshes and vector fields. We shall however refer to \cite{crane2018discrete} for a more comprehensive presentation and justification of the discrete differential approximations being used here. Let us assume that $q$ is a triangulated oriented surface mesh given by the ordered list of vertices $V=(v_1,v_2,\ldots,v_{N})$ with each $v_i\in\R^3$ and set of triangle faces $F$ where each $f\in F$ corresponds to an ordered triplet of distinct indices $f=(f_0,f_1,f_2)$ of $\{1,\ldots,n\}$. We then view the vector field $h$ as a list of vectors $(h_{i})_{i=1,\ldots,N}$ attached to each vertex of $q$. Note that, equivalently, one can interpret the discrete $q$ and $h$ as piecewise affine linear maps on each face of the mesh, by interpolation of the values at the vertices.  

We start with the $L^2$ term of the metric: $\int_{\pspace} \langle h,h \rangle \vol_q$. The discrete volume form can be first expressed over the mesh triangular faces. Specifically, for each face $f \in F$, we can calculate its area as $\vol_f = \|(v_{f_1}-v_{f_0}) \times (v_{f_2}-v_{f_0})\|$. The volume form on the vertices is then obtained by distributing the areas of the adjacent faces, namely for each vertex $v_i$, we take $\vol_{v_i} = \frac{1}{3} \sum_{f\ni i} \vol_{f}$. This leads to the following discrete version of the $L^2$ term:
\begin{equation*}
  \int_{\pspace} \langle h,h \rangle \vol_q \approx \sum_{i=1}^{N} \|h_i\|^2 \vol_{x_i}.
\end{equation*}

Next, we consider the first order terms of the metric. For any face $f \in F$, we can view both $q$ and $h$ as affine maps on $f$, by interpolation of their values at the three vertices of $f$. Then their differentials are constant on $f$ and given by the following $(3 \times 2)$ matrices: 
$$
dq_f = \begin{bmatrix}h_{f_1}-h_{f_0}, h_{f_2}-h_{f_0}\end{bmatrix} , \  dh_f=\begin{bmatrix}h_{f_1}-h_{f_0}, h_{f_2}-h_{f_0}\end{bmatrix} 
$$
We further have the following discrete versions of the metric $g_q$ and unit normal $n_q$ on the face $f$:
\begin{align*}g_f&=\begin{bmatrix}
\|e_{01}\|^2&e_{01}\cdot e_{02}\\
e_{01}\cdot e_{02}&\|e_{02}\|^2
\end{bmatrix},\\
n_f&=\frac{e_{01}\times e_{02}}{\|e_{01}\times e_{02}\|}.
\end{align*}
where $e_{01} = v_{f_1}-v_{f_0}$, $e_{02} = v_{f_2}-v_{f_0}$ are the two edges of the face $f$ passing through the vertex $v_{f_0}$. We then rely on the interpretation and the discretization of the different first order terms introduced in \cite{su2020shape}. Namely,
$$
\int_{\pspace} g_q^{-1}(dh_m,dh_m) \vol_q \approx \sum_{f \in F} \text{tr}(g_f^{-1} \delta g_f g_f^{-1} \delta g_f) \vol_f
$$
in which $\delta g_f$ represents the variation of the metric tensor $g_f$ resulting from the variation of the vertices of the mesh in the direction of the vector field $h$. In practice, in the computation of geodesics, $\delta g_f$ is calculated from one discrete time point to the next by taking the difference of the respective metric tensors of face $f$. Similarly,
$$
\int_{\pspace} g_q^{-1}(dh_+,dh_+) \vol_q \approx \sum_{f \in F} \text{tr}(g_f^{-1} \delta g_f)^2 \vol_f
$$
where each term inside the sum can be interpreted as the change of in the area of the face $f$ and
$$
\int_{\pspace} g_q^{-1}(dh_\bot,dh_\bot) \vol_q \approx \sum_{f \in F} \langle \delta n_f , \delta n_f \rangle \vol_f
$$
in which $\delta n_f$ stands for the variation of the normal vector $n_f$ resulting from the variation of the vertices of the mesh in the direction of the vector field $h$. The last first order term is discretized as follows:
$$
\int_{\pspace} g_q^{-1}(dh_0,dh_0) \vol_q \approx \sum_{f \in F} \text{tr}(g_f^{-1}\xi_f g_f^{-1} \xi_f^T) \vol_f
$$
where $\xi_f = dq_f^T dh_f - dh_f^T dq_f$.

\begin{figure}
\centering
\includegraphics[width=0.25\textwidth]{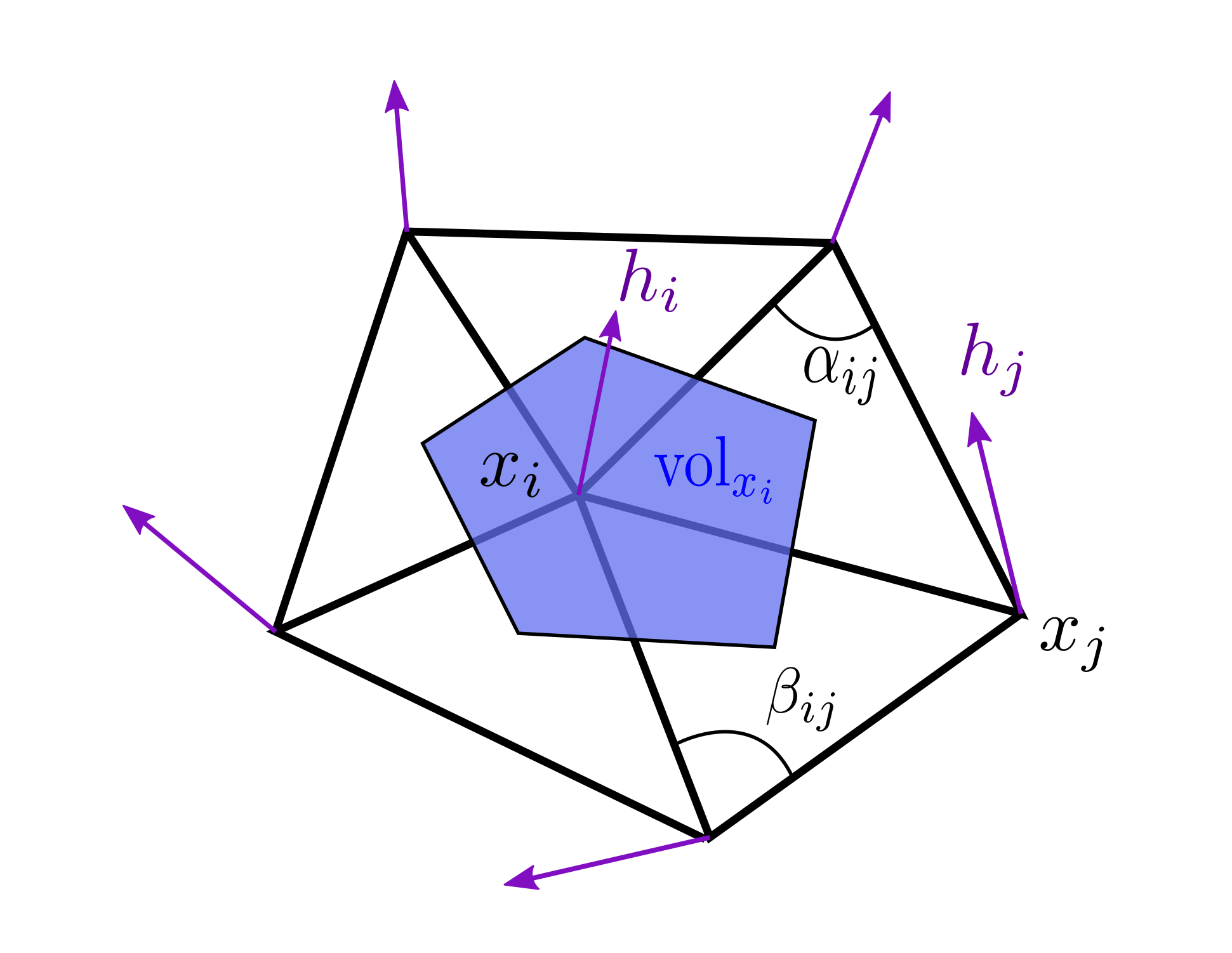}
\caption{Discrete volume form and Laplacian on a mesh.}
\label{fig:Laplace_discretization}
\end{figure}

Finally, for the discretization of the Laplacian in the second order term of the metric, we use the standard approximation on triangular meshes based on the cotangent formula. Letting $E$ be the set of oriented edges in the mesh viewed as ordered pairs of distinct vertex indices, we take for any $i \in \{1,\ldots,N\}$:
\begin{equation*}
    (\Delta_q h)_{v_i}=\sum\limits_{\substack{j|(i,j)\in E \\\text{ or }(j,i)\in E}}(\cot(\alpha_{ij})+\cot(\beta_{ij}))(h_i-h_j).
\end{equation*}
where $\alpha_{ij}$ and $\beta_{ij}$ are the angles defined as in Figure \ref{fig:Laplace_discretization}. Then the full discrete second order order term is obtained as:
$$
\int_{\pspace} \langle\Delta_q h,\Delta_q k\rangle \vol_q \approx \sum_{i=1}^{N} \| (\Delta_q h)_{v_i} \|^2 \vol_{x_i}.
$$

\section{Mesh invariant similarity measures}
In this section, we add some details regarding the similarity metrics being used in the registration procedure as well as for the evaluation and comparison of the different methods. With similar notations to the previous section, we consider two discrete surfaces $q$ and $q'$ with possibly different number of vertices and mesh structure. We denote by $(v_1,\ldots,v_{N})$ the vertices of $q$ and $F$ its set of faces, and similarly $(v'_1,\ldots,v'_{N'})$ and $F'$ the vertices and faces of $q'$.

First, we remind that the Hausdorff distance between the two shapes is given by the formula:
\begin{align*}
d_{\text{H}}(q, q') = \max\bigg\{&\sup_{i=1,\ldots,N} \inf_{j=1,\ldots,N'} \|v_i-v'_j\|,\\ 
&\sup_{j=1,\ldots,N'}\inf_{i=1,\ldots,N} \|v'_j-v_i\|\bigg\}
\end{align*}
In our numerical experiments, we use the approximate implementation provided by libigl~\cite{libigl}. Note that this metric is typically very sensitive to outliers.

In contrast, the Chamfer distance~\cite{fan2017point, groueix20183d} provides a more regular similarity cost which is defined as:
\begin{align*}
d_{\text{Ch}}(q,q') = &\frac{1}{N}\sum_{i=1}^N \inf_{j=1,\ldots,N'} \|v_i-v'_j\| \\  
&+ \frac{1}{N'}\sum_{j=1}^{N'} \inf_{i=1,\ldots,N} \|v'_j-v_i\|.
\end{align*} 
We use the Pytorch implementation of Thibault Groueix\footnote{\url{https://github.com/ThibaultGROUEIX/ChamferDistancePytorch}}. One of the downsides of this metric for comparing discrete surfaces, however, is that it is not necessarily robust to local changes of point density since it is designed as a distance between point clouds (without taking the triangle mesh into account) and it remains somewhat sensitive to outliers and noise \cite{Tong2021Balanced}. 

\begin{figure}
\centering
\begin{tabular}{ccc}
\includegraphics[width=0.15\textwidth]{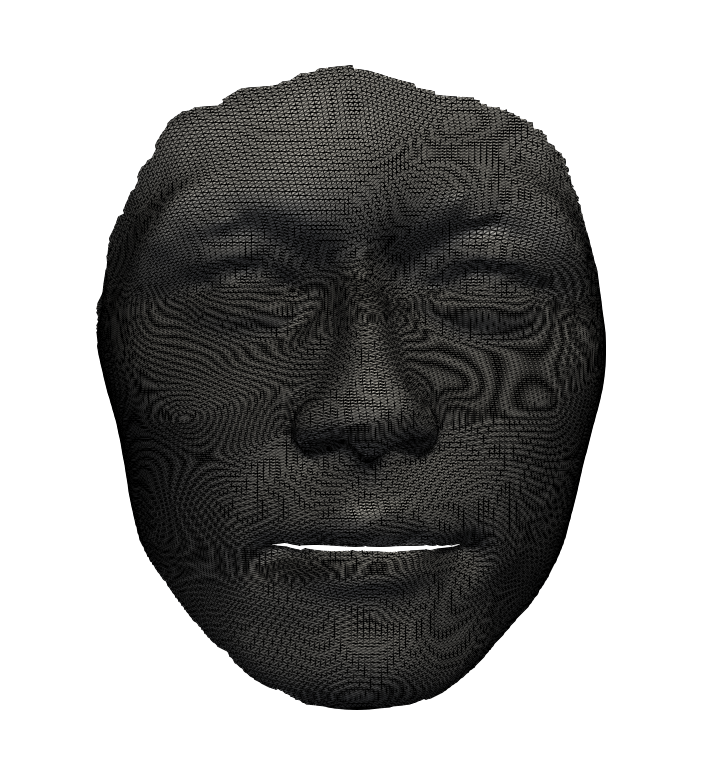} & & \includegraphics[width=0.15\textwidth]{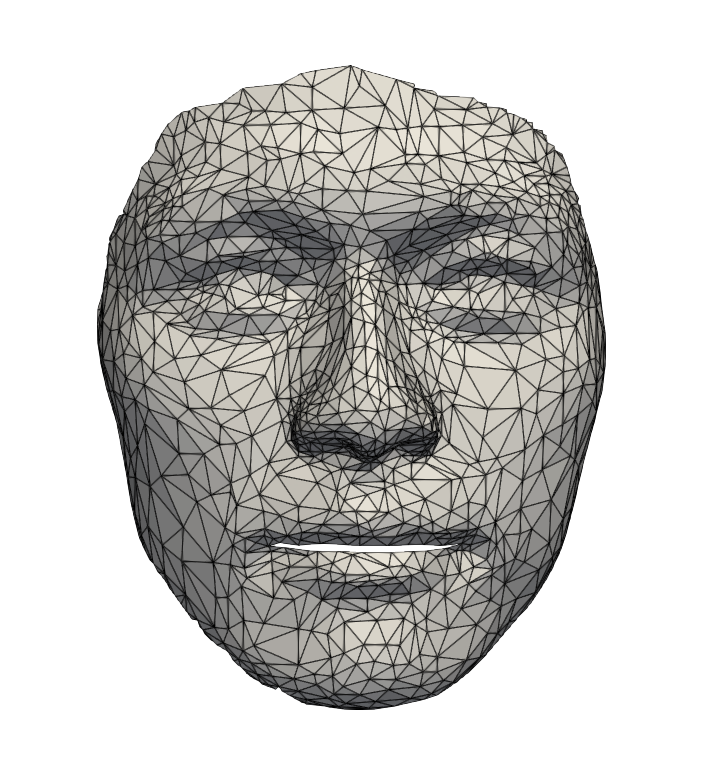} 
\end{tabular}
\begin{tabular}{c}
\includegraphics[width=0.25\textwidth]{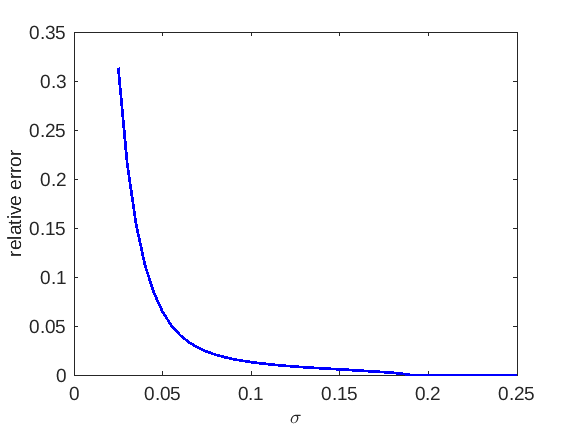}
\end{tabular}
\caption{Empirical illustration of the varifold distances approximate invariance to mesh sampling. Top row: a triangular mesh of a human face with 57,836 faces (left) and its downsampled version with 2000 faces (right). Bottom plot: the relative error in varifold norm $d_{Var}(q,q')/\|q'\|_{Var}$ between the full surface and the downsampled one, as a function of the kernel scale $\sigma$. One can see that this relative error remains close to $0$ for scales larger than $\sigma = 0.1$ but increases for smaller kernels. Note for reference that the surface diameter is normalized to $1$ while the average diameter of the mesh triangles in the original and downsampled mesh are respectively $9.4\times 10^{-3}$ and $4.6 \times 10^{-2}$.}
\label{fig:var_mesh_invariance}
\end{figure}

As similarity terms for the algorithms of this paper and final measure of reconstruction quality, we instead favor distances that are based on measure representations of shapes, as introduced in~\cite{charon2013varifold,kaltenmark2017general}. Specifically, we rely on the representation of surfaces as \textit{varifolds} equipped with kernel Hilbert metrics. The resulting family of metrics is equally defined for continuous and discrete surfaces and the properties of those metrics have been well studied, c.f. the aforementioned papers. In practice, they are computed via the following formula:
\begin{align*}
 &d_{\text{Var}}(q,q')^2 = \sum_{f,\tilde{f} \in F} k(c_f,n_f,c_{\tilde{f}},n_{\tilde{f}}) \vol_{f} \vol_{\tilde{f}} \\
 &-2\sum_{\substack{f \in F \\ f'\in F'}} k(c_f,n_f,c_{f'},n_{f'}) \vol_{f} \vol_{f'}  \\
 &+ \sum_{f',\tilde{f}' \in F'} k(c'_{f'},n'_{f'},c'_{\tilde{f}'},n_{\tilde{f}'}) \vol_{f} \vol_{\tilde{f}'}  \label{eq:varifold}
\end{align*}
where $c_f$ (resp. $c'_{f'}$) denote the barycenter of the triangle face $f$ (resp. $f'$) in $q$ (resp. $q'$). Here $k$ is a positive definite kernel function on $\R^3 \times S^2$. While several different families of kernels are possible (see discussion in~\cite{kaltenmark2017general}),  in all the experiments of this paper, we specifically take $k(x,n,x',n') = e^{-\frac{\|x-x'\|^2}{\sigma^2}}(n\cdot n')^2$ where $\sigma$ can be interpreted as a spatial scale of sensitivity of the metric which is chosen to be quite small ($\sigma=0.025$) in our examples. In this work, we adapted the Python implementation used in $H2\_SurfaceMatch$\footnote{\url{https://github.com/emmanuel-hartman/H2_SurfaceMatch}} which itself relies on the \textit{PyKeops} library \cite{feydy2020fast} for efficient evaluation and automatic differentiation of kernel functions on the GPU. We emphasize that such varifold metrics derive from distances between continuous surfaces which are independent of their parametrization. In practice, when considering discrete surface meshes, this typically leads to those metrics being approximately insensitive to variations in mesh sampling, at least for a certain range of kernel scale $\sigma$. We illustrate this property empirically with the example of Figure \ref{fig:var_mesh_invariance}.

 %We acknowledge that the varifold distance is specifically optimized in our approach, as such we display our results in terms of all three metrics.
%\begin{itemize}
%    \item Hausdorff distance between scan and matched shape
%    \item Difference of the integrals of mean curvature between the matched shape and the scan (Dirichlet Normal energy)
%    \item \textcolor{gray}{Varifold L2-distance between the scan and the matched shape }
%\end{itemize}

\section{Computational cost}
As stated in the paper, our pipelines are optimization based. We provide a substantial comparison for the different approaches. 
\begin{table}[]
    \centering
    \begin{tabular}{l|c|c|c|c}
    Method & Training & Retrieval & \multicolumn{2}{c}{Interpolation} \\ \hline \hline 
    LIMP & 1.5w & $<$1s & \multicolumn{2}{c}{$<$1s} \\ \hline
    3D-Coded & 12h & 160s & $<$1s & 160s \\\hline
    ARAPReg & 2w & 160s & $<$1s & 160s \\\hline
    BaRe-ESA & $<$1h & 160s & 91s & 160s \\
    \end{tabular}
    \caption{Computation costs for different methods. For the interpolation, the results are as follow: we display on the left the costs in the case latent codes are available, and the cost in the case they're not.}
    \label{tab:cost_train}
\end{table}

All the other approaches require significant training costs compared to BaRe-ESA which requires less than one hour, cf~\Cref{tab:cost_train}. 
%(more than a week for ARAPReg, Transmatching and LIMP, 12 hours for 3d-Coded, less than an hour for BaRe-ESA).
On the other hand, BaRe-ESA,  ARAPReg and 3d-Coded require additional optimization  for the latent code retrieval, which we found takes approximately the same time for all three methods. The optimization cost is driven by the mesh invariant costs -- varifold or Chamfer -- which have $n^2$ complexity, where $n$ is the number of vertices. LIMP is the only method that doesn't require optimization, but the network behaves notably bad when the poses are unseen as showed in the experiments. For the interpolation problem our method requires approximately 90 seconds if the latent codes are already available, whereas it takes approximately the same time as one latent code retrieval if they are not available.
All timing results were obtained using a standard home PC with a Intel 3.2 GHz CPU and a GeForce GTX 2070 1620 MHz GPU.% (of the order of 160s per latent code, except fr Transmatching for which it is 18s). For the interpolation problem our method requires approximately 90 seconds if the latent codes are already available, whereas it takes approximately the same time as one latent code retrieval if they are not available. All timing results, which will be added to the final version, were obtained using a standard home PC with a Intel 3.2 GHz CPU and a GeForce GTX 2070 1620 MHz GPU.

\section{Description of state-of-the-art methods}
We propose a detailed description of the state-of-the-art method we use as baselines. We selected deep learning methods that builds a flat latent space for human shape deformations. They describe as follows:
\begin{itemize}
    \item Learning Latent Shape Representations with Metric Preservation (LIMP) is a deep learning method modeling deformations of shapes using a variational auto encoder with geodesic constraints. The encoder part use a PointNet architecture, which makes it invariant to parameterization. The decoder part is a Multi Layer Perceptron. The geometric constraints are used a loss functions during the training process. The latent space is divided in an extrinsic part and an intrinsic part and the loss is applied on the interpolation in those dimensions. The intrinsic part is constrained using the computation of full geodesic matrix, which make the training process particularly heavy.
    \item As-Rigid-As-Possible Regularization (ARAP) is a deep learning method modeling deformations of shapes using an auto-decoder architecture. The latent codes and the decoder are learned altogether. During the training, an As-Rigid-As-Possible loss is imposed such that the decoder directions are similar to the ARAP ones. This procedure also makes the training procedure heavy. In order to make it parameterization invariant, we replace the $L^2$ metric by the varifold distance, as an alternative to our Riemannian latent space.
    \item 3D correspondences by deep deformation (3D Coded) is a deep learning method modeling deformations of shapes using a variational auto encoder. Similarly to LIMP, the encoder part use a PointNet architecture, which makes it invariant to parameterization. The decoder uses a Multi Layer Perceptron to deform a template mesh, but no constraint is imposed on the interpolation of latent variables. By taking advantage of a high number of training samples ($>200000$), they obtained state-of-the-art results for human shape correspondence.
    \item Functional Automatic Registration Method for 3D Human Bodies (FARM) is a functional-maps based approach for human body registration. The approach consists of multiple stages that enhance the initial mesh structure of a human body scan to propose a valid final functional map, based on a set of 15 landmark extracted automatically from the scan, between a given human body scan and a human body template. A final step of registration between the SMPL body model and the obtained correspondence is proposed.
\end{itemize}
In the paper, all those methods are trained using the same training set as Bare-ESA, from Dynamic FAUST and reported parameters from the respective papers. 

\section{Comparison to the framework of~\cite{hartman2023elastic}}
\begin{figure}
\centering
\includegraphics[width=\linewidth]{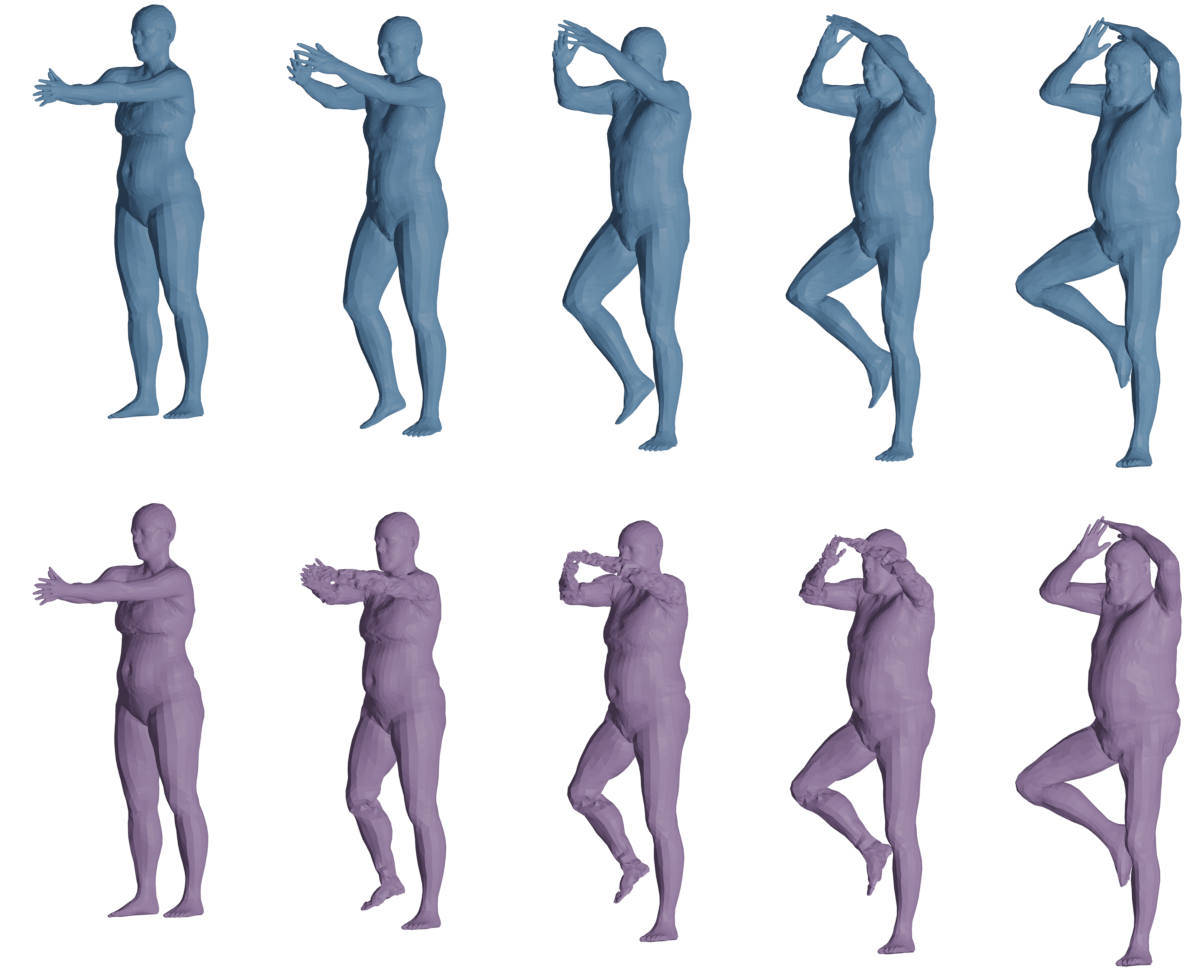}
\caption{First line: optimal deformation calculated using the basis informed ESA of the present article. Second line: optimal deformation calculated using a standard $H^2$-matching.}
\label{fig:basis_vs_nobasis}
\end{figure}

In Figure~\ref{fig:basis_vs_nobasis} we compare BaRe-ESA to the unrestricted method of \cite{hartman2023elastic}. Note, that BaRE-ESA is significantly cheaper to compute as we reduced the dimension of the minimization problem -- the latent space dimension will be in the order of 100s, while the dimension of the unrestricted method is on the order of 10000s. More importantly, one can observe that BaRe-ESA leads to significantly more natural deformations, cf. the movement of the arms in Fig.~\ref{fig:basis_vs_nobasis}.

\end{document}